\documentclass[journal]{IEEEtran}
\usepackage{cite}
\usepackage{amsmath,amssymb,amsfonts}
\usepackage{algorithm}
\usepackage{algorithmic}
\usepackage{graphicx}
\usepackage{epstopdf}
\usepackage{textcomp}
\usepackage{rotating}
\usepackage{booktabs}
\usepackage{caption}
\usepackage{color}

\usepackage{textcomp}
\usepackage{subfigure}
\usepackage{comment}
\newcommand{\markedred}[1]{\textcolor{black}{#1}}

\begin{document}

\title{DeepCount: Crowd Counting with WiFi via \\ Deep Learning}

\author{\IEEEauthorblockN{Shangqing Liu\IEEEauthorrefmark{1}\IEEEauthorrefmark{4},
Yanchao Zhao\IEEEauthorrefmark{1}\IEEEauthorrefmark{2}\IEEEauthorrefmark{4}, Fanggang Xue\IEEEauthorrefmark{1}, Bing Chen\IEEEauthorrefmark{1} and Xiang Chen\IEEEauthorrefmark{3} \\}
\IEEEauthorblockA{
\IEEEauthorrefmark{1} College of Computer Science and Technology, Nanjing University of Aeronautics and Astronautics, China\\
\IEEEauthorrefmark{2}Collaborative Innovation Center of Novel Software Technology and Industrialization, China \\
\IEEEauthorrefmark{3}Department of Computer Science and Technology, Nantong University, China \\
\IEEEauthorrefmark{4}Co-First Authors
}
}



\maketitle

\begin{abstract}
Recently, the research of wireless sensing has achieved more intelligent results, and the intelligent sensing of human location and activity can be realized by means of WiFi devices. However, most of the current human environment perception work is limited to a single person's environment, because the environment in which multiple people exist is more complicated than the environment in which a single person exists. In order to solve the problem of human behavior perception in a multi-human environment, we first proposed a solution to achieve crowd counting (inferred population) using deep learning in a closed environment with WIFI signals - DeepCout, which is the first in a multi-human environment. step. Since the use of WiFi to directly count the crowd is too complicated, we use deep learning to solve this problem, use Convolutional Neural Network(CNN) to automatically extract the relationship between the number of people and the channel, and use Long Short Term Memory(LSTM) to resolve the dependencies of number of people and Channel State Information(CSI) . To overcome the massive labelled data required by deep learning method, we add an online learning mechanism to determine whether or not someone is entering/leaving the room by activity recognition model, so as to correct the deep learning model in the fine-tune stage, which, in turn, reduces the required training data and make our method evolving over time. The system of DeepCount is performed and evaluated on the commercial WiFi devices. By massive training samples, our end-to-end learning approach can achieve an average of \markedred{86.4\%} prediction accuracy in an environment of up to 5 people. Meanwhile, by the amendment mechanism of the activity recognition model to judge door switch to get the variance of crowd to amend deep learning predicted results, the accuracy is up to \markedred{90\%}.

\end{abstract}


\IEEEpeerreviewmaketitle

\section{Introduction}
The WiFi signals has spanned over the city and reveal the sensing ability such as activity recognition, human identification, localization and beyond. However, current application of WiFi sensing is only effective in single person scenario which greatly restricts the use of WiFi environmental sensing \cite{ref4}\cite{ref7}\cite{ref11}\cite{ref9}. Research of indoor crowd counting is the basis of multi-object environmental sensing and various potential applications, e.g. tour guiding and crowd control. Meanwhile, crowd-counting in WiFi environment is a very challenging task, as the WiFi signal is very arbitrary owing to the uncertainty of the states in the room. Thus, traditional signal process and pattern recognition methods for activity recognition \cite{ref5}\cite{ref29}\cite{ref30} are powerless in extracting the information from multiple overlapping signals to perform crowd-counting.

Traditionally, the most popular crowd-counting are based on computer vision techniques with images from camera \cite{ref1}\cite{ref21}\cite{ref22}\cite{ref23}. Recently, electromagnetic waves method are implemented with special hardware \cite{ref2}\cite{ref24}. However, camera based approaches may suffer from blind spots in the corner or absence of light conditions. It also introduce privacy issues. Special hardware system such as WiTrack\cite{ref2} mainly utilize TOF (Time-of-Flight) by FMCW (Frequency Modulated Continuous Wave) to provide delegate and well-constructed signal. However, these devices suffers from high deployment cost and thus not comparable to ubiquitous WiFi deployment. Some researchers also used smart phones\cite{ref12,ref25,ref26} to infer the number of speakers in a dialog. However, these methods are not device-free and not friendly for the elderly and children. Based on these, we believe that the WiFi-based crowd counting study is very meaningful and it can solve the above problems well.

In this paper, our proposed DeepCount solution will have different different multi-path distortions and unique pattern of waveform in WiFi signals based on human activity. The physical amplitude and phase information will be distorted greatly owing to human activities in the WiFi environment during the process of WiFi signal propagation. Hence, we can utilize this time series of Channel State Information (CSI) values captured in the WiFi signals for sensing. Compared to existing crowd counting system using Wifi eFrogEye\cite{ref17}, which predict the number of people based on the utilizes Grey model, the model could not fully utilize the phase and amplitude information. Thus, we utilize both two dimensions and the powerful deep learning method to realize such function. 

However, to realize this idea into a working system, we face a variety of technical challenges. The first technical challenge is how to extract feature values that can  simulate the relationship between CSI values and population counts.After a detailed analysis of a large amount of CSI data, we found that traditional features such as entropy, maximum or variance do not meet the demand. Due to the uncertainty of the state of the people in the room, we cannot find the correlation and pattern through simple mathematical modeling. In this case, we need to preserve the original features of the data as much as possible during data processing, so we only perform simple denoising on the amplitude and phase information.

The second technical challenge is extract the counting model from complex overlapping signals. Traditional training methods such as Support Vector Machine (SVM), Bayesian Classifier cannot capture the overlapping features and the background noise. Recent advancement of deep learning and the success of its application in computer vision has shed the light on resolving our problem. As the extremely complicated relationships between CSI waveforms and crowd counting, we utilize the neural networks including Convolutional Neural Networks (CNNs) and Recurrent Neural Networks (RNNs) with Long Short Term Memory Network (LSTM) to construct this complicated model.

The third challenge is the massive data required to build the model with deep learning and how to build a robust model with proper structure of neural networks. Regarding the data acquisition, the training process can easily lead to overfitting with inadequate or biased data. Thus, we design massive and proper designed data process to ensure the quality and quantity. Specifically, we collect 6 different activities up to 10 people, which cover different indoor behaviors such as walking, talking and different participants. Then, to improve the diversity, we split the data in different time window length and label them properly. Regarding the proper network structure for our problem, basically, we use CNNs and LSTMs to for the deep neural network model. In addition, regularization and exponential decay methods are applied to avoid overfitting. 

The fourth challenge is how to adapt our deep learning model over time. Although the deep learning method has a strong learning ability, the learned model could loss over time for slight environment change. To overcome this, our basic idea is using our activity recognition model under the condition of the single person entering/leaving the room. Then, we can infer a state change for current model. At the same time, if the deep learning model gives an error result which means the increment/descendent of the number of people in the room greater than 1 compared with the previous slot, we can correct this result to finetune the parameters of the last layer of the neural network in our deep learning model. With this mechanism, we will eventually improve the accuracy of WiFi counting model up to \markedred{90\%} in a relatively
robust manner.

Our method fully utilizes the characteristics of amplitude and phase of CSI with a specifically designed CNN-LSTM network, where the CNN is applied to extract the deep features while LSTM is applied to handle time series signal.  Meanwhile, to make our model more flexible to adapt the time evolving of the crowd-counting in an online manner, we add an online learning mechanism to correct our deep learning model by fine-tune the last layer parameters of neural network. Such method endow our method with time-evolving features, thus greatly improve the practicality. 


The main contributions of this paper can be summarized as follows:
\begin{itemize}
\item We theoretically analyze the correlation between crowd counting and the variation of CSI and utilize deep learning to characteristic this relationship. To the best of our knowledge, this is the first solution to solve the population of WiFi signal populations using neural networks. It proposes a new approach to solving such problems.
\item We proposed the \textit{DeepCount} system and adopted a deep learning approach to solve multi-person context awareness problems. We use LSTM and CNN to implement automatic extraction of features.
Then we use a softmax layer for crowd counting.
\item To further improve the performance of DeepCount, we add an online learning mechanism by our activity recognition model, the experiments show that by this mechanism, we can eventually get the \markedred{90\%} accuracy. 
\item We introduced some simple and effective denoising methods that eliminate noise while maximizing the characteristics of the data.
\end{itemize}

The remaining of this paper is organized as follows.Section \ref{relatedwork} provides a background about neural network and related work of WiFi sensing. We will analyse the characteristics about CSI and the reasons to choose deep learning to solve this problem in Section \ref{understanding}. The details in our design of DeepCount will be discussed in Section \ref{system design}. The implementation and evaluation are presented in Section \ref{evaluation} followed by conclusions in Section \ref{conclusion}.

\section{BACKGROUND and RELATED WORK}\label{relatedwork}

\subsection{Deep Neural Networks}

With the rapid development of artificial intelligence, Recurrent Neural Networks (RNNs) and Convolutional Neural Networks (CNNs) have demonstrated impressive advances in numerous tasks.

\textbf{Recurrent Neural Networks:}
RNNS are designed for processing sequential information due to the ability of preserving the previous information to the current state. RNNs loop the same network each time step to capture the state information and use this information with the next input as the whole input information. However, RNNs suffer from the vanishing gradient problem that leads to failure in learning long sequences. LSTM was motivated to overcome the limitation by introducing a new structure called memory cell that has additional “forge” gates over the simple RNNs. Its unique mechanism enables it to capture long term dependencies. Hence, RNNs with LSTMs have been widely used in processing long sequences.

\textbf{Convolutional Neural Networks:}
CNNs have achieved so many the state of the art works in the field of computer vision and NLP recently for the ability of automatically high-level features extraction. The traditional block of CNNs contains three parts: a convolutional layer, activation function and pooling layer. The main function of convolution layer is to extract features automatically with a filter. The filter is a small square with common shape (3,3) or (5,5), which serves as dot product among dimensions of input data. An activation function such as Sigmoid, Relu follows every convolution layer to perform non-linear transform. Pooling layer is used to reduce dimension of input data but retain the most important information. There are different types of pooling such as Max and Average. In case of max pooling, it extracts the max value in the predefined area. The above three parts are the basic blocks of CNNs.

\subsection{WiFi Sensing}
WiFi sensing can be applied in the fields of activity recognition, indoor localization and user authentication. Here we discuss briefly about the related works. 

\textbf{Activity Recognition:} In the single environment, many researchers have utilized CSI values in large scale activity\cite{ref27}\cite{ref29}\cite{ref30} and small scale motion recognition\cite{ref28}\cite{ref31}\cite{ref32}. For example, in the terms of large scale activity recognition, WiFall\cite{ref3} detected the activity of fall with an accuracy of 87\% using a novel detection method. E-eyes\cite{ref4} used CSI histograms as fingerprints to separate different daily activities. CARM\cite{ref5} proposed CSI-speed model and CSI-activity model to analyse the correlation between CSI values and human activities. WiDir\cite{ref7} used the physical Fresnel zone model to estimate the moving direction of humans. Meanwhile, in the terms of small scale motion recognition, WiFinger\cite{ref9} used CSI to recognize a set of gestures including (0-9) and WiHear\cite{ref11} used directional antennas to hear people talks by catching CSI data caused by lip movement.

\textbf{Indoor Localization:} Another important part in WiFi sensing is to use CSI for indoor localization \cite{ref33} \cite{ref34}. SpotFi\cite{ref14} utilized phase information from CSI values with music algorithm achieved decimeter level accuracy. Based on this, Li \textit{et al.} \cite{ref16} from Peking University improved the music algorithm and proposed dynamic-music algorithm to measure angle-of-arrival(AOA) of signals. LiFS\cite{ref6} localized a target without offline training with the help of "clean" subcarriers. Some researchers also used deep learning techniques with CSI for localization. DeepFi\cite{ref15} trained weights and biases between network layers as fingerprints for localization. 

\textbf{User Authentication:} We can use CSI for authentication and privacy protection for the reason that CSI contains the state information of wireless channel. Liu \textit{et al.} \cite{ref19} proposed a system to construct user profile resilient to the presence of spoofer. WiWho\cite{ref10} and WifiU\cite{ref8} used WiFi devices to analyse unique gait patterns for human identification. 

In recent work similar with ours, Xi \textit{et al.} \cite{ref17} proposed the system called FCC which uses WiFi signals to get the number of people in the indoor environment. However, FCC applied the Verhulst theory which was not sufficiently reliable and the algorithm of Dilatation-based crowd profiling could not accurately describe the correlation between CSI and crowd counting. Similarly, Domenico\cite{ref38} tried to find the correlation between the number of people and CSI. And the Euclidean distance of two CSI waveforms was used for identification, which was not powerful to depict the relationships between the number of people and CSI. Shi \textit{et al.} \cite{ref20} applied DNN(Deep Neural Network) for user authentication. The system achieved over 94\% and 91\% authentication accuracy with 11 subjects through walking and stationary activities respectively. Compared to these, in our training process, we use LSTM to capture long term dependencies and CNN to extract features automatically to improve the performances of neural network on the crowd counting. 

\begin{figure*}[htb]
\begin{minipage}[t]{0.24\textwidth}
\centering 
\includegraphics[width=1\textwidth]{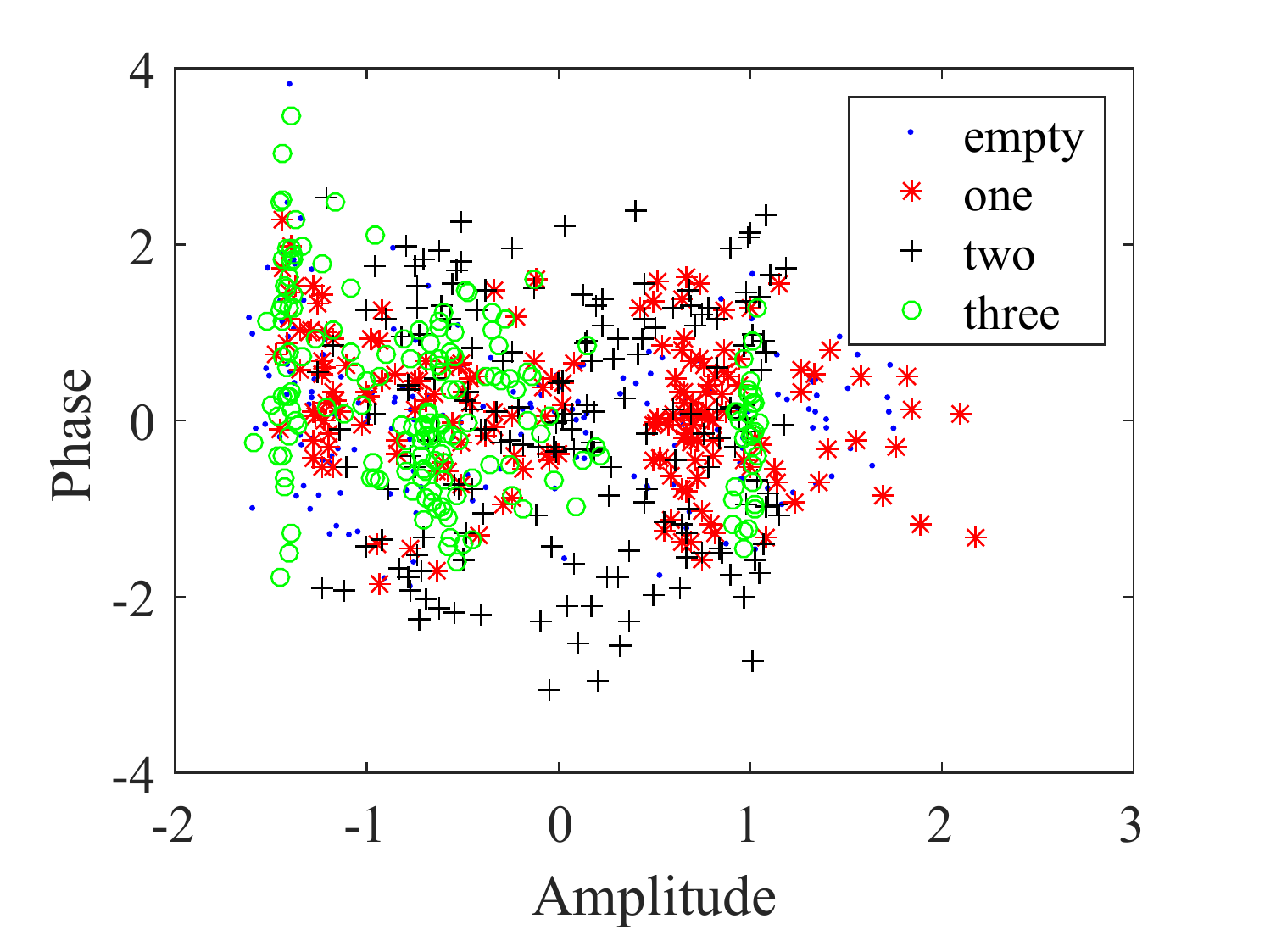} 
\caption{Raw normalized CSI amplitude and phase information}
\label{raw}
\end{minipage} 
\begin{minipage}[t]{0.24\textwidth} 
\centering 
\includegraphics[width=1\textwidth]{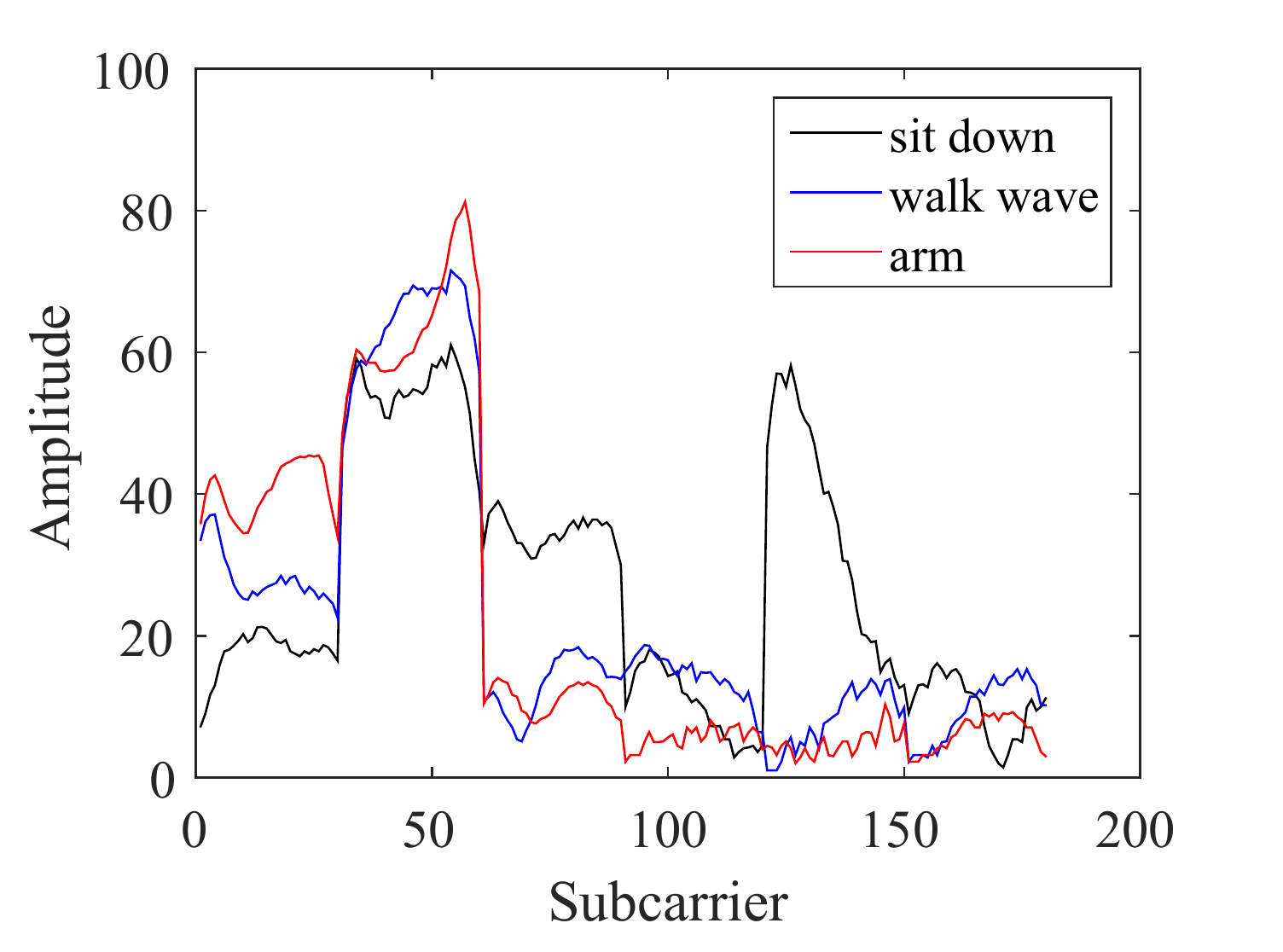} 
\caption{Amplitudes of different subcarriers at different activities}
\label{different}
\end{minipage}  
\begin{minipage}[t]{0.24\textwidth} 
\centering 
\includegraphics[width=1\textwidth]{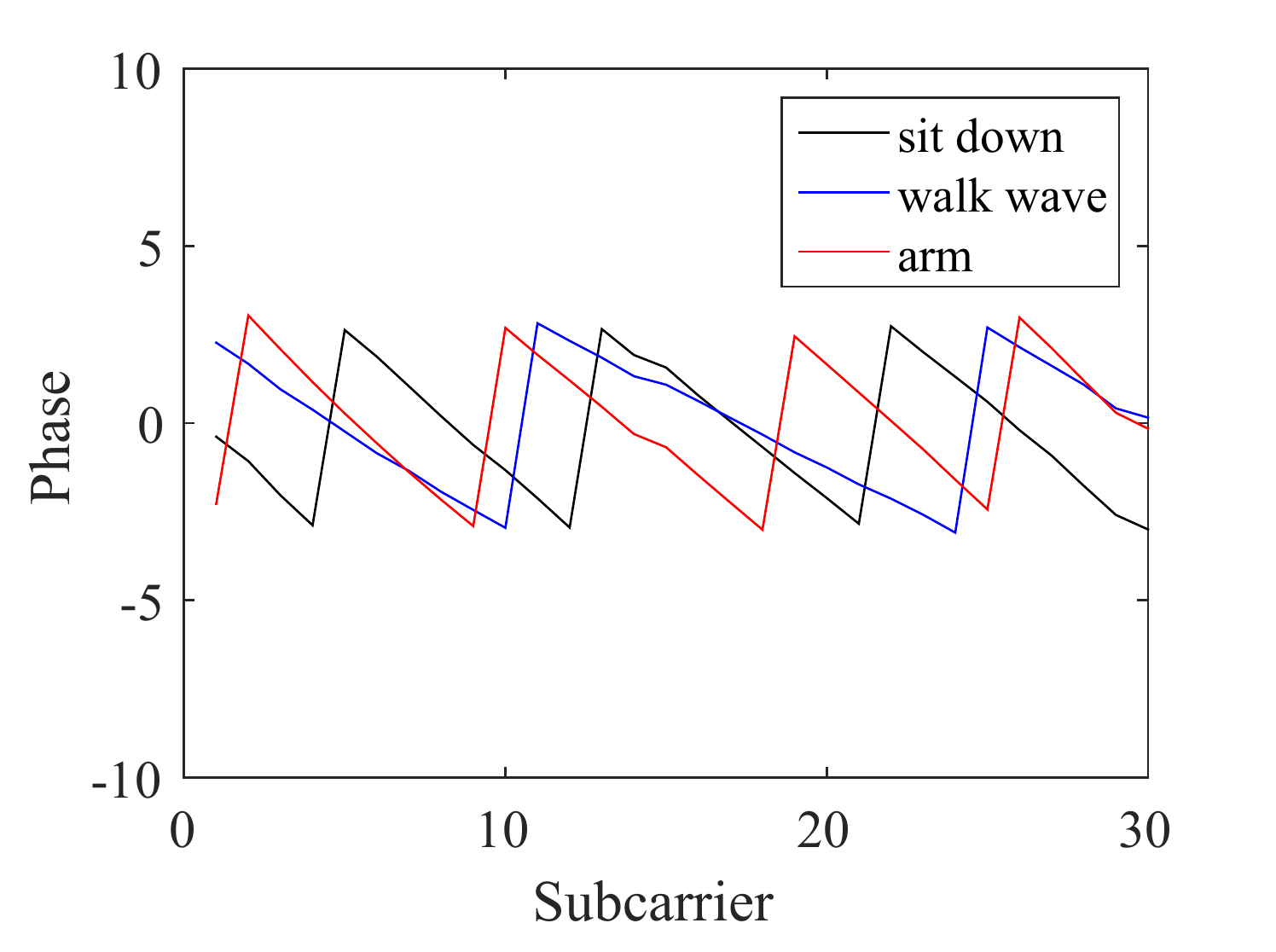} 
\caption{Phases of different subcarriers at different activities}
\label{phase}
\end{minipage}  
\begin{minipage}[t]{0.24\textwidth} 
\centering 
\includegraphics[width=1\textwidth]{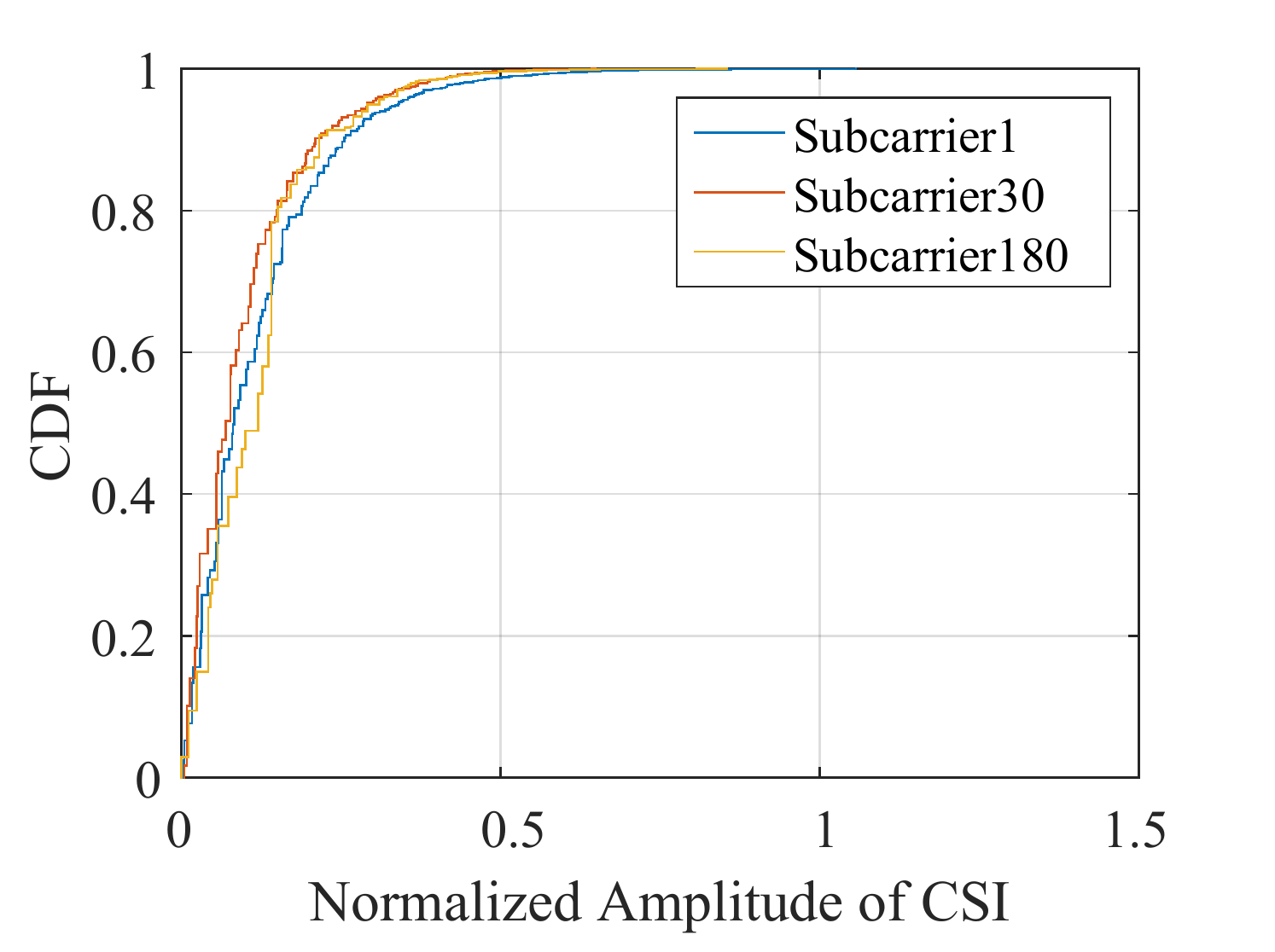}  
\caption{CDF of normal amplitudes of CSI values at a fixed activity}
\label{same}
\end{minipage} 
\end{figure*} 

\section{Analysis of WIFI COUNTING} \label{understanding}
\subsection{Overview of CSI}
In the wireless communication environment, Channel State Information (CSI)  represents the nature of the wireless link, which describes how the WiFi signal propagates from the transmitter to the receiver, with the combined effects of scatter, distance-varying fading. Today's WiFi devices are usually equipped with multiple transmitting and multiple receiving antennas for high data transmitting, which are called MIMO (Multiple Input, Multiple Output).The physical link between the pair of transmitting and receiving antennas has multiple subcarriers at the same time. Hence, at the time point $t$, we can get enough data. WiFi signals can be expressed as: 
\begin{equation}
Y(t)=H(t)X(t)+N
\end{equation}
where  $Y(t)$ and $X(t)$ are the  receiving signals and  transmitting, $H(t)$ is the Channel Frequency Response(CFR) , $N$ is noise vector. Furthermore, CFR can be expressed as:
\begin{equation}
H(t)=\sum_{k=1}^Na_k(t)e^{-j2s\pi\tau_k(t)}
\end{equation}
where $a_k(t)$ denotes the channel attenuation and initial phase offset and $e^{-j2s\pi\tau_k(t)}$ denotes the phase offset caused by the propagation delay. CSI is the estimation of CFR, which contains both amplitude and phase information. The amplitude represents the strength of WiFi signals and phase information represents the periodic variation of the signal with the propagation distance. In the physical level, phase changes one period, the signal propagates the length of one wavelength. Hence, the amplitude and phase are sufficient for us to sense environmental changes.

\subsection{WiFi Counting Model}\label{WiFi Counting Model}
From the prior analysis, we can easily find WiFi sensing has three kinds of applications: activity recognition, indoor localization and human authentication. Since different human activities have different effects on WiFi signals, some simple machine learning methods can separate these known activities based on CSI. However, using a WiFi signal to detect the number of people in a room is much more complicated because the state of the person in the room is unknown. We were unable to separate the WiFi signal into the specified activity and further determine the number of people in the room. In other words, we are unable to extract features that are directly related to the number of people manually.

Since we can't manually extract features, can we look for another way to solve this problem? May be we can dig some inspirations from computer vision. The researchers in computer vision use deep learning approaches \cite{ref35}\cite{ref36} to extract features automatically to get the state-of-art performances. In some specific tasks, these applications even exceed the performances of humans. Hence, we solve this problem based on neural networks. We have three reasons to believe that deep learning can solve this problem.
\subsubsection{Neural network can effectively solve nonlinear classification problems}
In a traditional linear model, the output is the linear weighted sum of the inputs. The linear model can be expressed as:
\begin{equation}
\label{equation}
y=\sum_{i}w_ix_i+b
\end{equation} 
where ${w_i}$, ${b}$ are the parameters of the model. The equation \ref{equation}  is a linear transformation that can be used to solve linear classification problems. However, for nonlinear classification problems, the equation \ref{equation} does not meet the requirements. Therefore, we need to use a multi-layered neuron with activation to deal with this type of problem. From Fig.\ref{raw}, we find that the relationships between crowd counting and amplitude and phase information are non-linear. In theory, if the hidden layer contains enough neurons, the neural network can fit any complex nonlinear function.  Hence, if we use proper neural network architecture, we may get better results.

\subsubsection{CSI values exhibit large variability among different subcarriers}
In our experiment, a laptop with three antennas as a receiver and one with two antennas as a transmitter. Each WiFi link between transmitter and receiver has 30 subcarriers. Hence totally we get 180 subcarriers. From Fig.\ref{different}, we can find CSI amplitude at a specific activity has different variations among subcarriers and different activities show differences on the signals. Similar property also shown in the Fig. \ref{phase}. Hence, based on these, amplitude and phase information are sufficient as input data and we can use LSTMs to capture the long term dependencies during activity segments and CNNs to get features automatically.
\subsubsection{CSI values are stable at a fixed subcarrier}
Although CSI values exhibit differently among subcarriers, these values also show great stability over time at the same subcarrier, which proves the model we build may be robust. Fig.\ref{same} plots the CDF of the standard deviation of CSI amplitudes. 80\% of the standard deviations are below 20\% of the average value at a fixed subcarrier, which depicts that CSI values are much more stable compared with RSSI values.
\begin{figure}[htb]
\centering 
\begin{minipage}[t]{0.3\textwidth}
\includegraphics[width=1.2\textwidth]{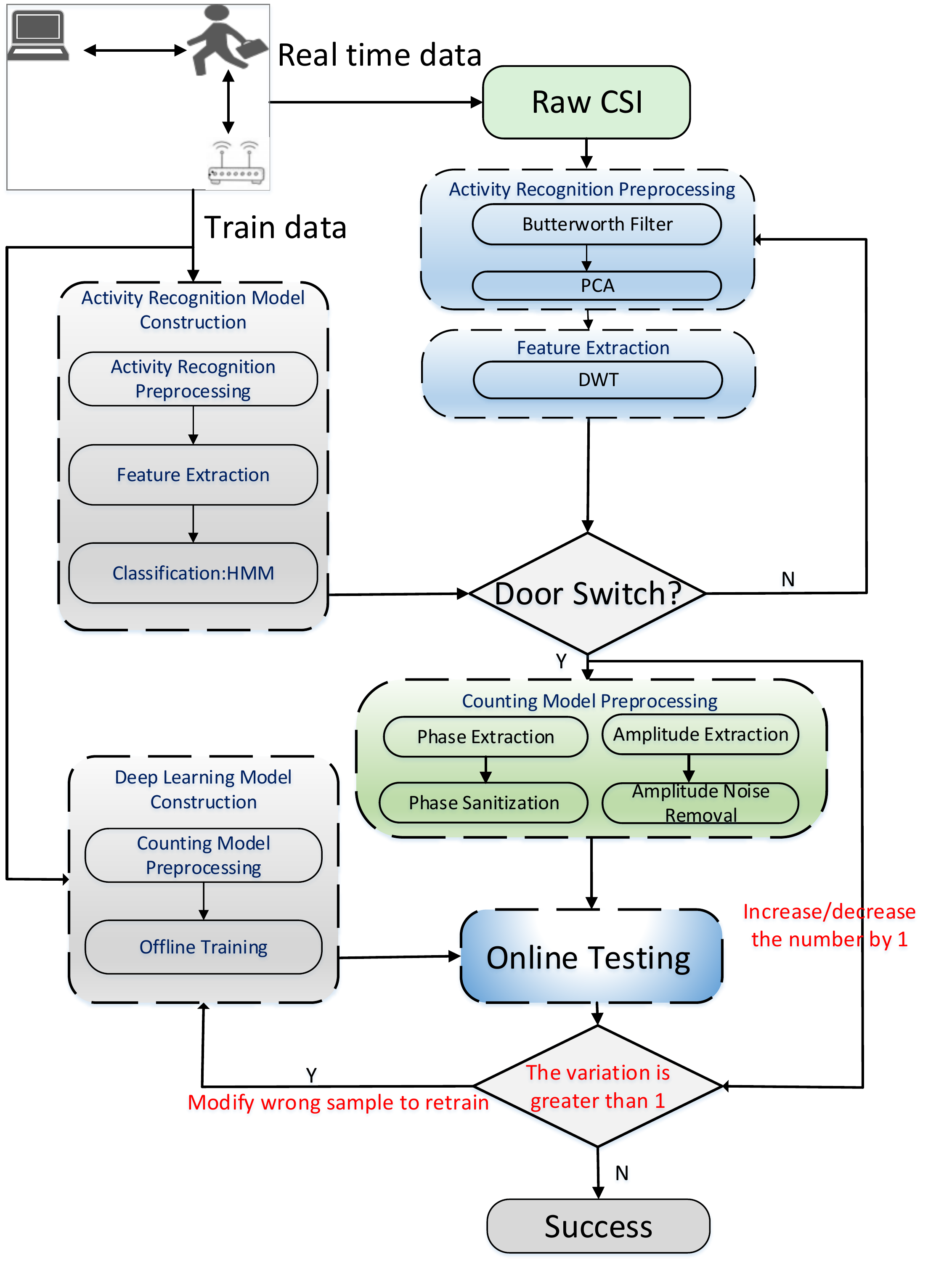} 
\caption{Framework of DeepCount}
\label{framework}
\end{minipage} 
\end{figure}

\section{SYSTEM DESIGN}\label{system design}
\subsection{DeepCount Overview}
Our DeepCount system uses the neural network to automatically learn the relationship between crowd counting and CSI based on  that human activities have a significant impact on WiFi signals. From Fig.\ref{framework} we can find that there are two modules of our DeepCount system : activity recognition model and deep learning model. DeepCount uses a router with two transmit antennas to transmit signals and a laptop with three receive antennas to receive signals. Hence, our data contains 180 streams.We can extract phase and amplitude information from each stream. For Activity recognition model under the condition of single person entering/leaving the room, we only use the amplitude information and this model contains three parts: \textit{Activity Recognition Preprocessing, Feature Extraction and Classification}. For activity recognition Preprocessing. We apply the Butterworth and PCA to remove the noises and electromagnetic inference in the environment. Then, we extracted the feature information from these amplitude information by DWT (Discrete Wavelet Transformation) and use these features to construct our activity recognition model with HMM (Hidden Markov Model). By this activity recognition model, we can distinguish the activity of door switch which means someone opening the door and entering into the room/closing the door and going out of the room among other activities. 
Our deep learning model contains three parts: \textit{Counting Model Preprocessing, Offline Training and Online Testing}.We use both phase and amplitude to participate in the preprocessing of the counting model, rather than some systems, such as electronic frog eye \cite{ref17}, which use amplitude or amplitude information alone. We need to eliminate the obvious noise in this information before using it for offline training for better results. First, we divide the data set into 3 levels. The Dataset-fixed represent activities and positions that are fixed. The Dataset-semi means we allow volunteers to conduct free activities on the premise of a fixed position, they are free to choose actions which may be a combination of fixed actions. The Dataset-open means volunteers can do anything anywhere in the room. Then DeepCount employs a deep neural network with CNNs and LSTMs to train the samples ,whit  a large number of weights and biases ,the deep neural network will  extract feature based fingerprints which can effectively represent the relationship between crowd counting and the CSI variances. In the online test stage, we used trained models to predict the number of people in the room. At the same time, we also monitor the activity of door switch among other human activities using activity recognition model. Once there is someone entering/leaving the room, the increment/decrement of the number of people is greater than 1 compared with the previous moment by our deep learning model, which means the predicted result is wrong. We label this sample and add this data to retrain the parameters of the last layer in our deep learning model. 

\subsection{CSI Collection}
When the transmitter continuously transmits the WIFI signal, the receiver continuously receives the WIFI signal at the same time, and the DeepCount will automatically extract the CSI value through the CSI tool provided on the receiver.  We fixed sampling rate is 1500 packets/s to ensure fine grained information about human activity. For each packet, the CSI matrix H is extracted. Owing to there aretwo antennas of the transmitter (Tx) and three antennas of receiver (Rx) , we can express the matrix H can as:
\begin{equation} 
\left( 
\begin{array}{ccccc}      
     H_{1,1} & H_{1,2} & \cdots & H_{1,30}\\ 
     H_{2,1} & H_{2,2} & \cdots & H_{2,30}\\ 
     \vdots & \vdots & \ddots & \vdots\\ 
     H_{6,1} & H_{6,2} & \cdots & H_{6,30}
     
\end{array} 
\right) 
\end{equation} 
 where $H_{i,j}$ represents the CSI values at $j_{th}$ subcarrier in the $i_{th}$ Tx-Rx pair. Hence, we can get three dimensional data $CSI=[H_1,H_2,...,H_t]$ for the duration time ${t}$. We extract the phase and amplitude values from the raw CSI data for further processing.

\begin{figure*}
 \subfigure[Raw CSI amplitude]{
 \label{Raw CSI Amplitude}
  \includegraphics[width=0.23\textwidth]{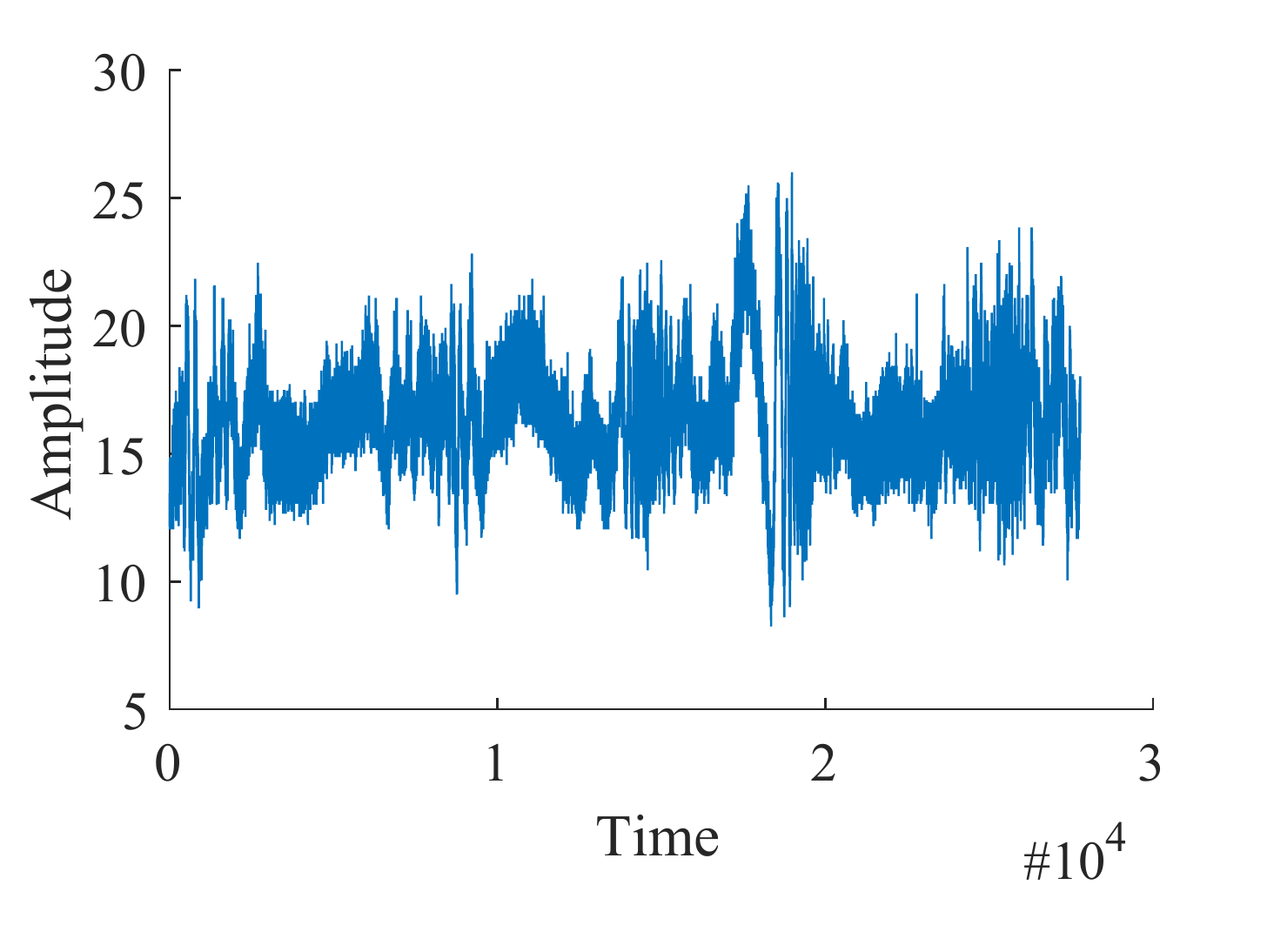}
  }
 \subfigure[Raw data after butterworth filter]{
 \label{Butter}
  \includegraphics[width=0.23\textwidth]{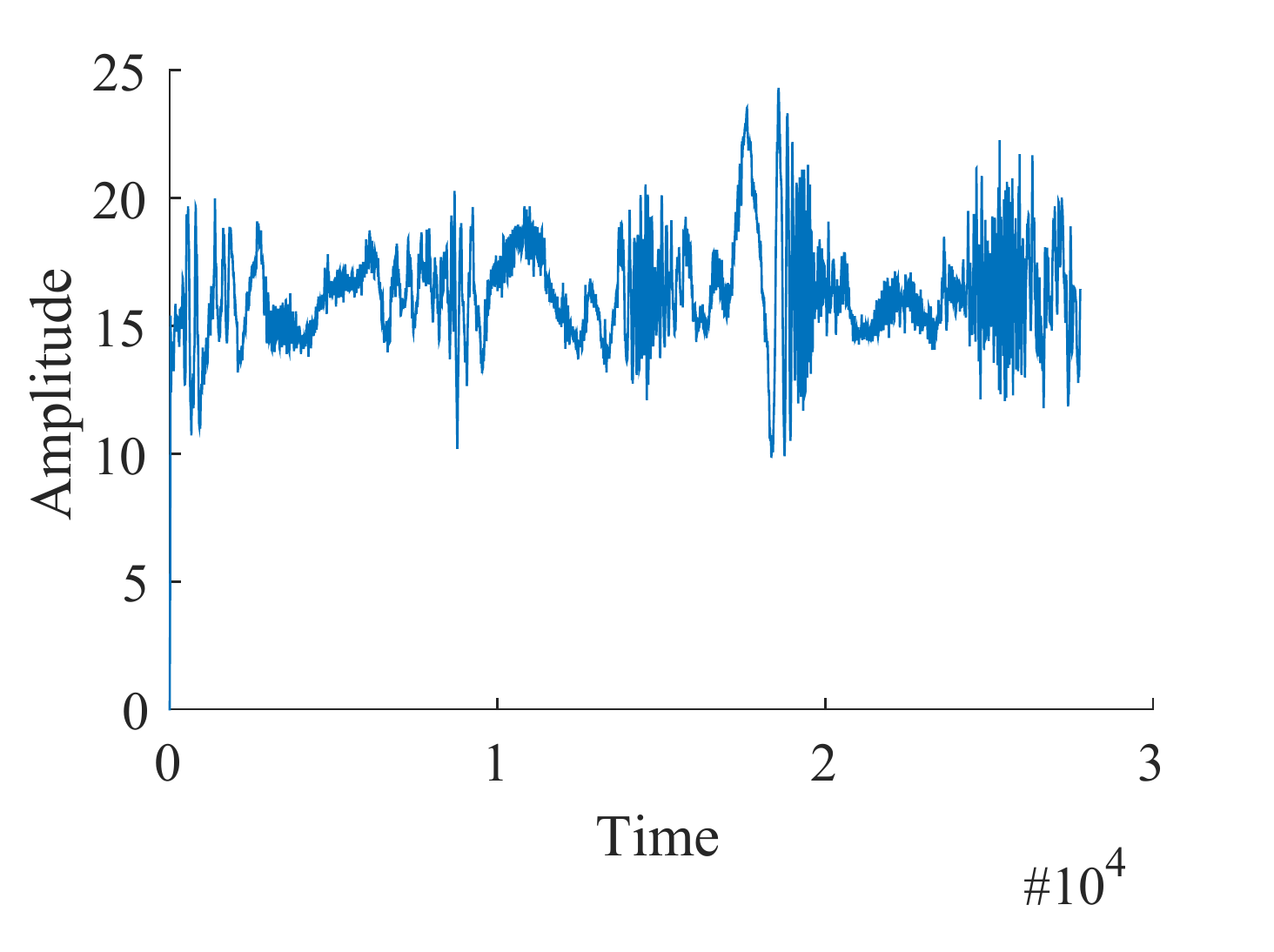}
  }
  \subfigure[Weighted moving average filter]{
  \label{Weight}
  \includegraphics[width=0.23\textwidth]{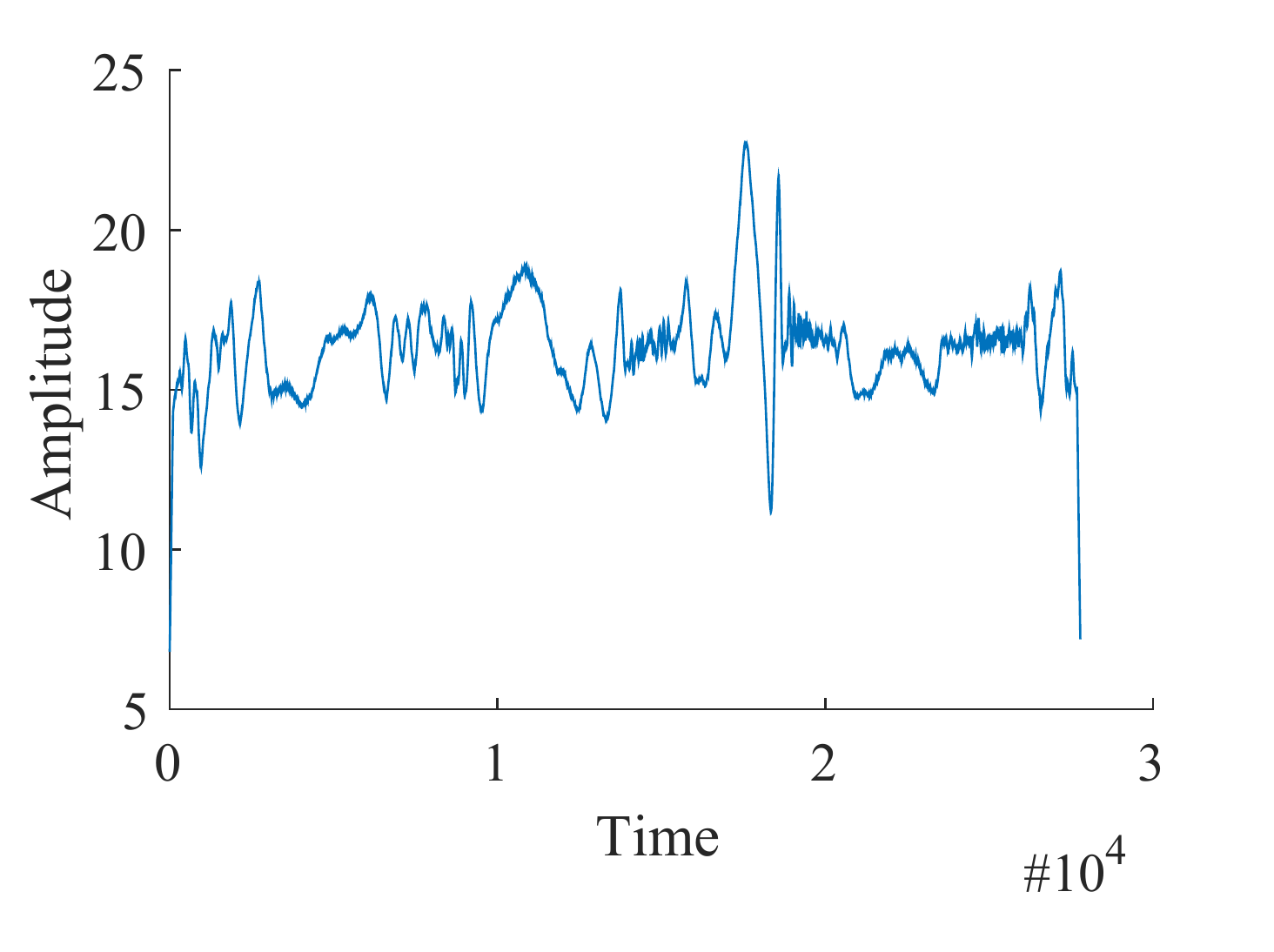}
   }
  \subfigure[Raw data after PCA]{
  \label{PCA}
  \includegraphics[width=0.23\textwidth]{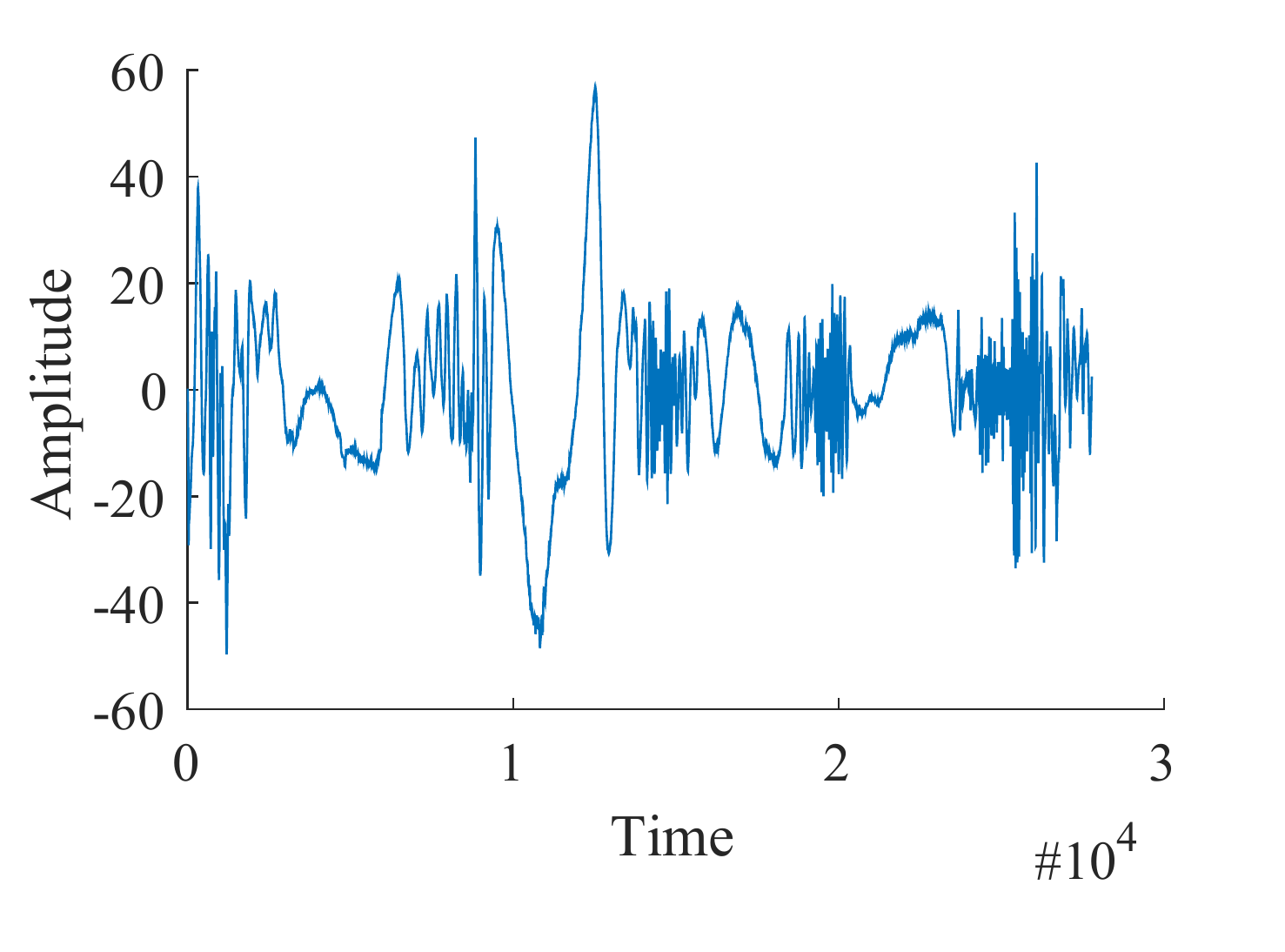}
   }
   \caption{CSI walking amplitude data at subcarrier 30}
\end{figure*}

\subsection{Activity Recognition Model Construction}
\subsubsection{Activity Recognition Preprocessing} \label{Activity Recognition Preprocessing}
As is shown in Fig. \ref{Raw CSI Amplitude}, raw CSI amplitude data contain too many noises. In our activity recognition model, the speeds of human activities such as sitting, walking, door switch are not very fast and the signal changes caused by these activities lie at a low frequency spectrum while the noises caused by hardware have a relatively high frequency spectrum. Hence, Butterworth low-pass filter is a natural choice which removes high frequency noises. Actually, frequency variations in CSI streams due to human normal activities are often within 200Hz, hence we set a cut-off frequency of 200Hz. The waveform after Butterworth filter is shown in Fig. \ref{Butter}, it is obvious to find that high frequency noises are removed. However, the noises between 1 and 200Hz cannot be eliminated, we utilize PCA to reduce the noises further. DeepCount applies PCA to CSI streams by the following three steps:
\begin{itemize}
 \item \textit{DC Component Removal:} In this step, we first subtract the corresponding constant offsets from CSI streams to remove the Direct Current (DC) component of every subcarrier. The constant offsets can be calculated through a long-term averaging over that subcarrier. 
 \item \textit{Principal Components:} DeepCount calculates the correlation matrix Z as $Z=H^{T} \times H$ and gets the eigenvector $q_{i}$ through eigen decomposition of Z. After that, DeepCount achieves the principal components by the following equation:
 \begin{equation}
h_{i}=H \times q_{i}
 \end{equation}
 where $q_{i}$ is the $i^{th}$ eigenvector and $h_{i}$ is the $i^{th}$ principal component.
 \item \textit{Smoothing:} Finally, we apply a 5-point median filter to avoid abrupt changes in the CSI streams which are likely to facilitate the results.
\end{itemize}
DeepCount discards the first principal component $h_1$ and retains the next ten principal components to be used for feature extraction. It is mainly due to following reasons. From a large number of experiments, we observed that noises are mainly captured in the first component. However, the information about human activities is captured in all principal components. Since the PCA components are uncorrelated, we can discard the first principal component without losing too much useful information. Fig. \ref{PCA} shows the third PCA component of our method and we find that the signal is much smoother.

\subsubsection{Feature Extraction}
The responses of different activities on the frequency spectrum are different. The CSI frequency is determined as $f = 2 \nu / \lambda$, where $\nu$ is the speed of human activity and $\lambda$ is the WiFi signal wavelength. From the equation, we can find that the frequencies of running and walking are obviously different due to the speed of running is much faster than walking normally. Hence, Discrete Wavelet Transformation (DWT) is a proper choice. The DWT employ functions that are both in time and frequency which overcome the limitations of the classical Fourier Transform and meanwhile, DWT provides more appropriate wavelet basis to choose. The discrete wavelet function is defined as:
\begin{equation}
\Phi_{m,n}(t)=a_0\phi(a_0^{-m}t-nb_0)
\end{equation} 
The discrete wavelet coefficient is defined as:
\begin{equation}
W_{m,n}(t)=\int_{- \infty}^{+ \infty} f(t)\overline{\Phi_{m,n}(t)}dt
\end{equation} 
In the DWT, the signal is decomposed into a coarse approximation coefficients and detail coefficients. Then the coarse approximation coefficients are further decomposed using the same wavelet function. In DeepCount, we chose the Daubechies D4 wavelet to decompose the PCA component into 10 levels of frequencies spanning 1 Hz to 200 Hz. Then, in order to capture the details of different activities, we use a time window of size 128 to average the detail coefficients. In each window, we use its average energy and variance as features. The corresponding feature matrix is ​​shown in the matrix below：
\begin{equation}
\left(
\begin{array}{ccccc}
     E_{1,1} & E_{1,2} & \cdots & E_{1,n-1} & E_{1,n}\\
     E_{2,1} & E_{2,2} & \cdots & E_{2,n-1} & E_{2,n}\\
     \vdots & \vdots & \ddots & \vdots & \vdots\\
     E_{10,1} & E_{10,2} & \cdots & E_{10,n-1} & E_{10,n}\\
     V_{1,1} & V_{1,2} & \cdots & V_{1,n-1} & V_{1,n}\\
     V_{2,1} & V_{2,2} & \cdots & V_{2,n-1} & V_{2,n}\\
     \vdots & \vdots & \ddots & \vdots & \vdots\\
     V_{10,1} & V_{10,2} & \cdots & V_{10,n-1} & V_{10,n}\\

\end{array}
\right)
\end{equation}
where $E_{i,j}$ is the average energy at the $i^{th}$ level in the $j^{th}$ time window, $V_{i,j}$ is the variance at the $i^{th}$ level in the $j^{th}$ time window and $n$ is the number of time windows.

\subsubsection{Classification}
After the above steps, we get the feature matrix representing different activities. We utilize Hidden Markov Model (HMM) to train the features for the reason that most of human activities can be divided into different phases, for example door switch can be divided into silent, acceleration, deceleration and stop four phases, which corresponding to the concept of states in HMM. Markov Model is a special kind of Bayesian network. The variable $Y_t$ denotes ${t^{th}}$ node in the network, each node has S possible states and different states have different transition probabilities. For the variables $Y_1 \cdots Y_T$, we have
\begin{equation}
P(Y_1 \cdots Y_T)=P(Y_T|Y_{T-1})P(Y_{T-1}|Y_{T-2})\cdots P(Y_2|Y_1)P(Y_1)
\end{equation}
The probability distribution of the state at ${t}$ moment depends only on the state at ${t-1}$, which is called transition probability. HMM is an extension of Markov Model which means the actual state at the moment t is unknown, instead we can only get the observation ${X_t}$, and the observation and state are not one-to-one. A state may have several kinds of observations and an observation can also have multiple states with different probabilities. Each time when a state is entered, a feature vector is genera probabilistic. HMM utilizes the transition probability which provides more information compared to the traditional training methods.

\subsubsection{Monitor with Activity Recognition Model}
We can easily find that the impacts on CSI spectrum from different activities are different by the analysis of \ref{WiFi Counting Model}. Hence, we utilize HMM to construct the activity recognition dataset with some daily activities such as walking, falling, running and the activity of door switch(entering/leaving the room) under the condition of single person entering/leaving the room. Then we can use this dataset directly to monitor the door switch activity in real time. The high accuracy of activity recognition we can find from Fig. \ref{activity} provides a judgment for deep learning model to correct its predicted result with previous result, which means that once the activity recognition model find there is someone entering into room, the increment should be equal to 1.

\subsection{Deep Learning Model Construction}
\subsubsection{Counting Model Preprocessing}
\begin{figure*}
 \subfigure[Raw wrapped CSI phase]{
 \label{Raw Wrapped CSI Phase}
  \includegraphics[width=0.33\textwidth]{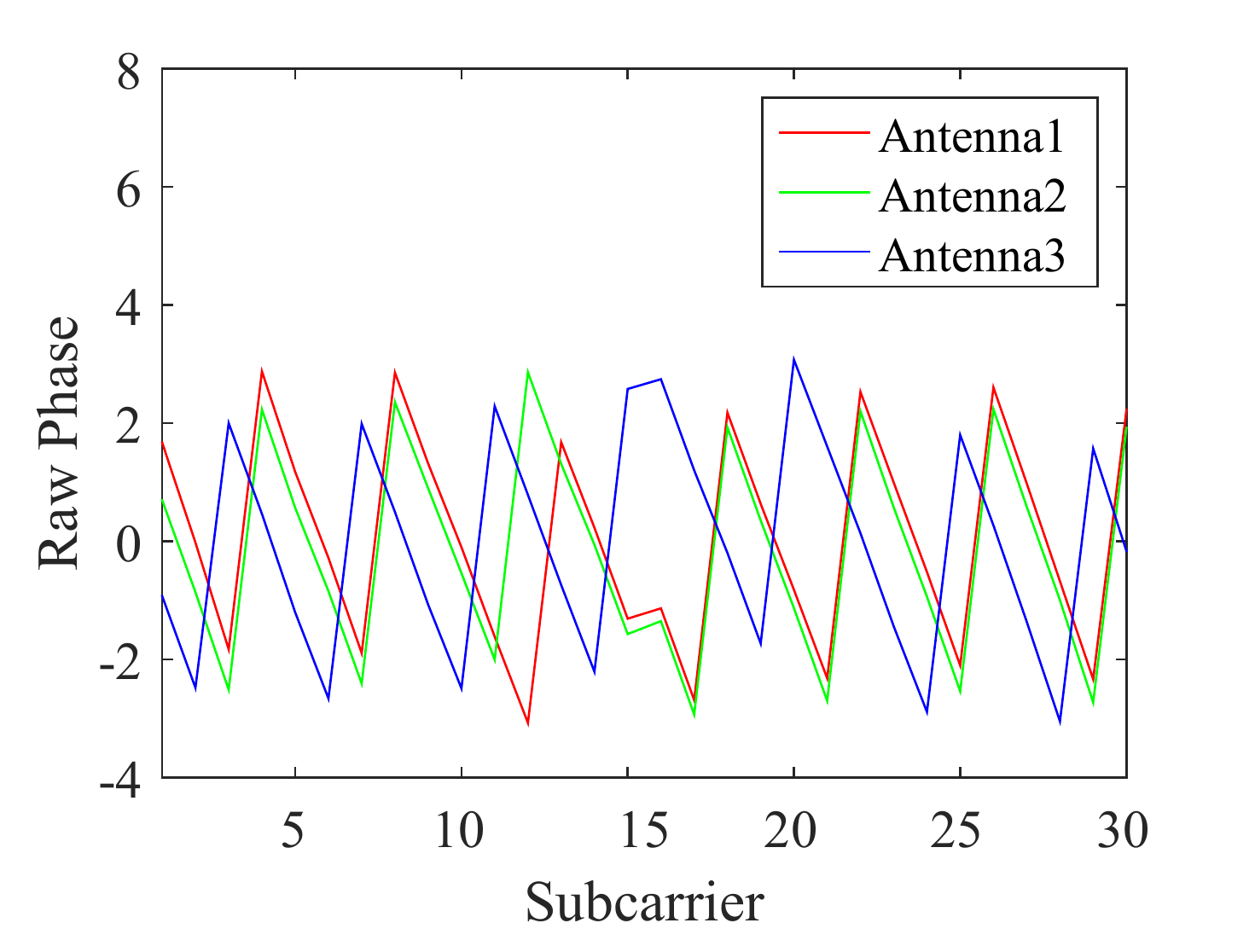}
  }
 \subfigure[Unwrapped CSI phase]{
 \label{Unwrapped CSI Phase}
  \includegraphics[width=0.33\textwidth]{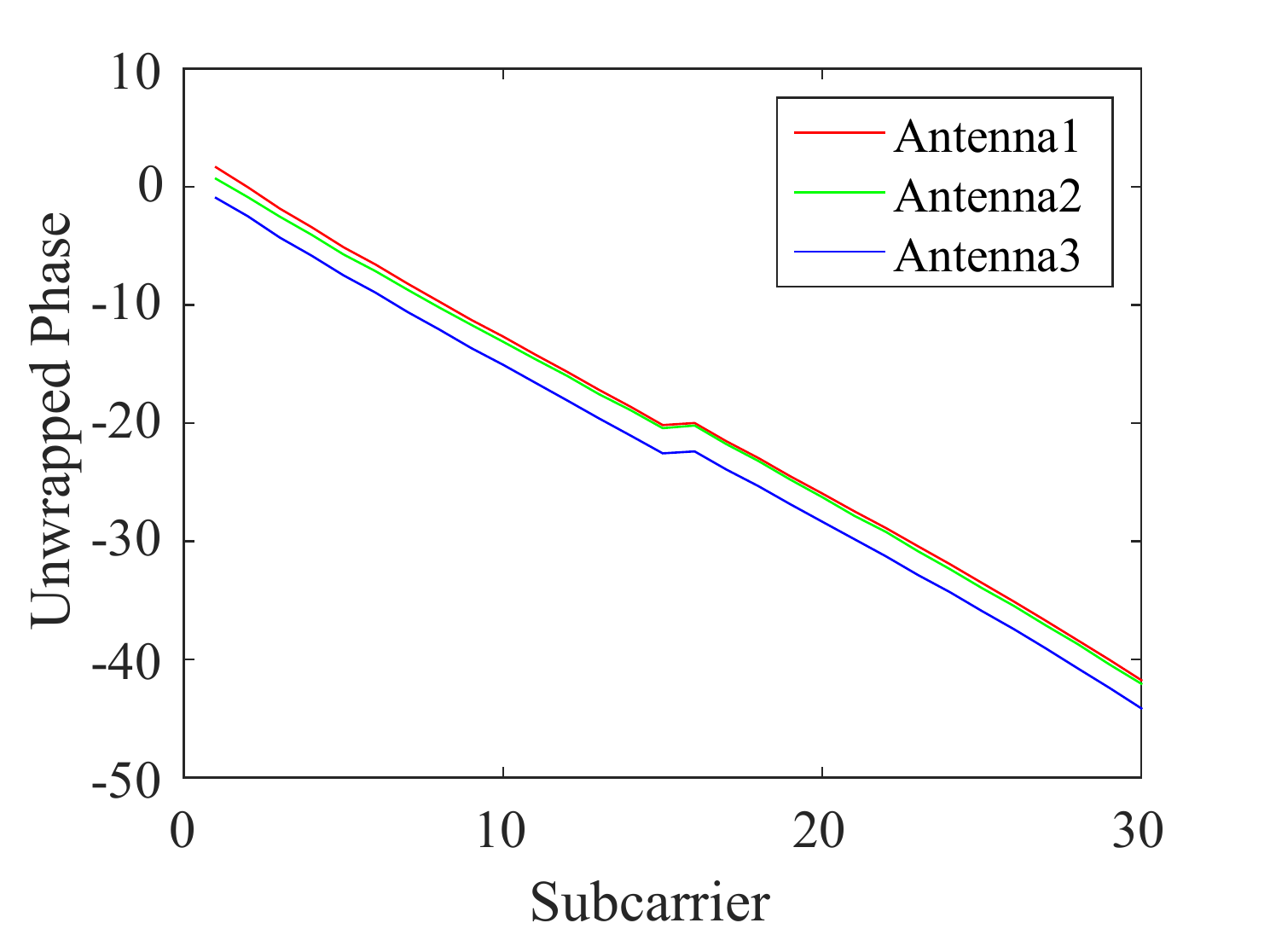}
  }
  \subfigure[Modified CSI phase]{
  \label{Modified CSI Phase}
  \includegraphics[width=0.33\textwidth]{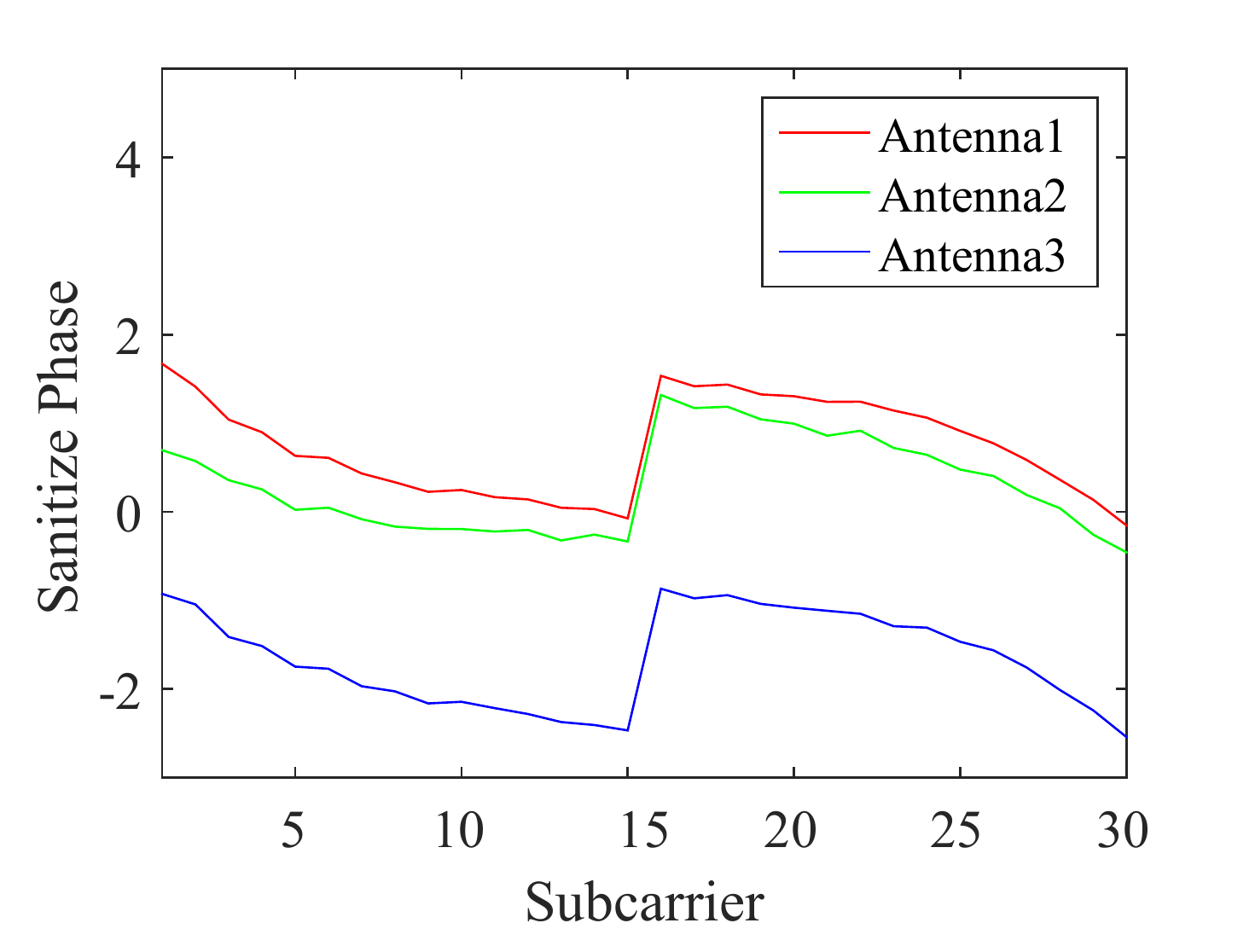}
   }
   \caption{Phase sanitization}
\end{figure*}

The counting model preprocessing is different from \ref{Activity Recognition Preprocessing} for the reason that We need to preserve the original feature of the signal as much as possible for classification. However, PCA denoising method loses original features of waveform shown in the Fig. \ref{PCA} which is harmful for deep learning classification. Hence, we apply the weighted moving average method to remove the noises of amplitude for deep learning model. At the same time, We add the phase information as features to improve the performance of deep learning model.
\begin{itemize}
\item{\textit{Amplitude Noise Removal:}}
We utilize the Weighted Moving Average（WMA） algorithm to remove noises. Specially, we use the equation \ref{weightedequation}  for the amplitude values of the $1^{th}$ subcarrier $Sub_1=[A_1,A_2,...,A_t]$ to get weighted averaged amplitude values:

\begin{equation}
\small
\label{weightedequation} 
A_t^{'}=\frac{1}{m+...+1} \cdot [m \cdot A_t+(m-1) \cdot A_{t-1}+...+A_{t-m+1}]
\end{equation}

for the time $t$ , $A_t^{'}$ is the weighted averaged value ， $m$ is the weighted relationship between current values and the historical values. We empirically fixed $m=100$  in our system. Fig.\ref{Weight} shows the results of weighted average filter, the waveform is much smoother compared to raw data.

\item{\textit{Phase Sanitization:}}
Although we get the phase data, we cannot use these data directly. Owing to the carrier frequency offset (CFO)\cite{ref37} and the sampling frequency offset (SFO), the phase $P_M$ we get can be expressed as:
\begin{equation} 
\label{phaseequation} 
P_M=P+2\pi\frac{m_i}{N}\Delta t+\beta+N
\end{equation}
where $P$ is the genuine phase, $\Delta t$ is the time lag due to SFO, $\beta$ is the unknown phase offset due to CFO and $N$ is the noise. From equation \ref{phaseequation}, we find that owing to the unknown $\Delta t$ and $\beta$, we cannot get the real phase. However, from the equation, we found that linear fit can eliminate the effects of SFO and CFO. Fig.\ref{Raw Wrapped CSI Phase} shows the raw phase $P_M$ values when someone is walking in the room. We can see that the initial phase values are folded within the range of $[-\pi,\pi]$ .  In order to get the true phase values, we unfolded the CSI phases which is shown at Fig.\ref{Unwrapped CSI Phase} firstly. Next, we need to remove the impacts of SFO and CFO. We get the mean phase values $y$ for antennas on each subcarrier. Then we utilize the linear fit to get the true phase values. The whole algorithm is shown in Algorithm \ref{algorithm}. Fig.\ref{Modified CSI Phase} presents the modified phase values by our algorithm.

\begin{algorithm}[h]
  \caption{Phase Sanitization}
  \begin{algorithmic}[1] 
  \label{algorithm}
  \REQUIRE ~~\\ 
  The raw phase values $P_M$;\\ 
  The number of subcarriers $Sub$;\\ 
  The number of Tx-Rx pairs $M$; 
  \ENSURE ~~\\ 
  The calibrated phase values $P_C$; 
  \FOR{$i=1$ to $M$}		           
    \STATE $U_P=unwrap(P_M(:,i))$;
  \ENDFOR
   \STATE $y=mean(P_M,2)$;
   \FOR{$i=1$ to $Sub$}		           
    \STATE $x=(0:Sub-1)$;
  \STATE $p=polyfit(x,y,1);$;
  \STATE $yf = p(1)*x;$;
  \FOR{$j=1$ to $M$}		           
    \STATE $P_C(:,j)=P_M(:,j)-yf$;
  \ENDFOR
  \ENDFOR
  \end{algorithmic}
\end{algorithm}
\end{itemize}

\subsubsection{Offline Training}
\begin{figure}[htb]
\begin{minipage}[t]{0.5\textwidth}
\centering 
\includegraphics[width=0.9\textwidth]{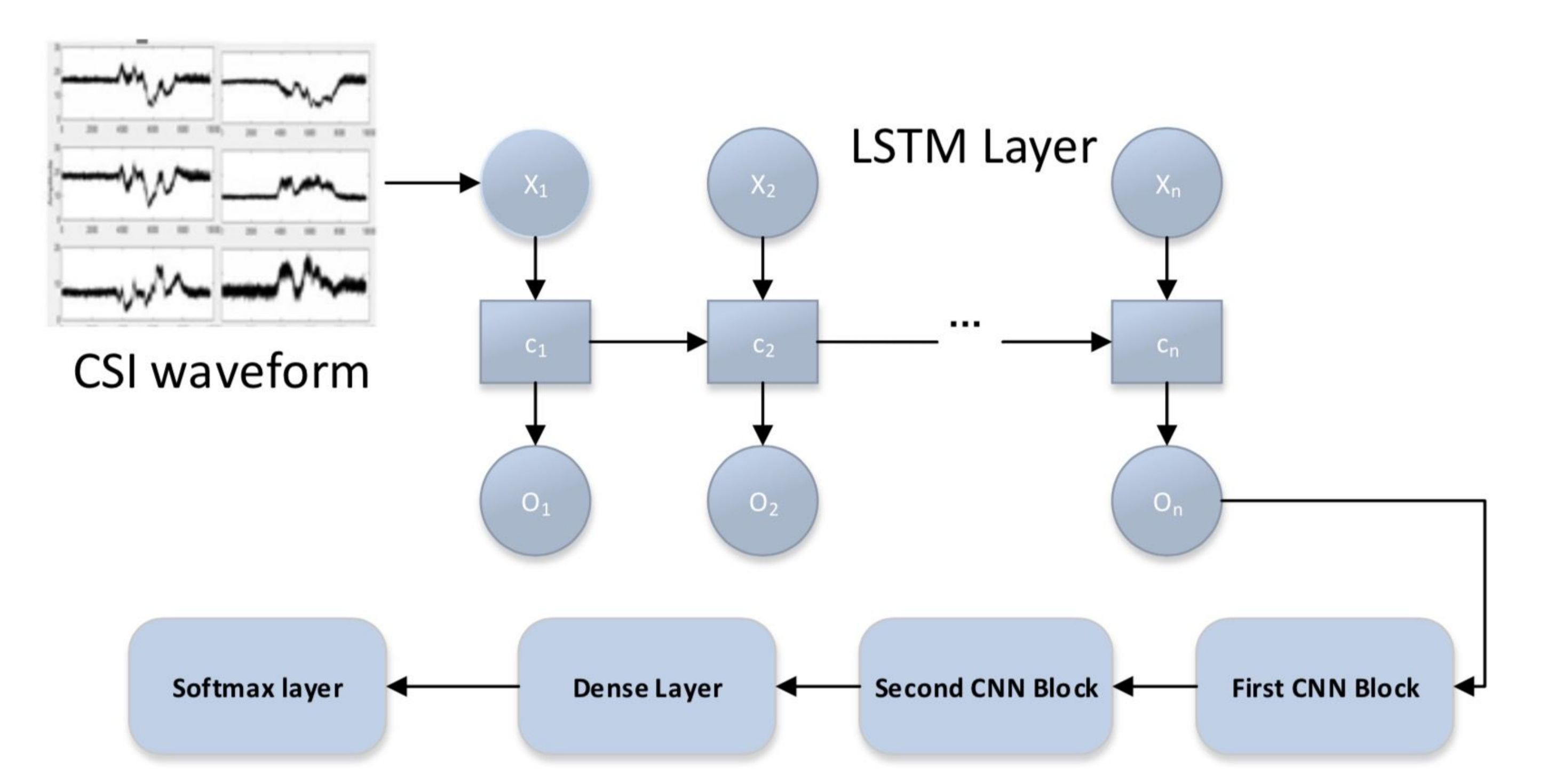} 
\caption{The architecture of network}
\label{network}
\end{minipage} 
\end{figure}

Fig.\ref{network} illustrates the proposed network architecture for Crowd Counting Training. This component comprises these four parts sequentially.

\begin{itemize}
\item{\textit{LSTM Layer:}}
We use a LSTM layer to extract long short dependencies of activity segments. After processing of the amplitude noise removal and phase sanitization, the CSI information of different activities are then passed to an LSTM layer to get long short dependencies. The output of LSTM is then input into convolutional layers for higher level features. Here we use one layer of LSTM. Because , for LSTM networks, one layer is powerful enough and much easier to tune hyper parameters. The input data has 360 dimensions including 180 dimensions for amplitude and 180 dimensions for phase information. Hence, the shape of input data is $time \times 360$. Let N be the number of units in the LSTM layer, the output of the layer for a single activity becomes a $ time \times N \times 1$ vector. In our experiment, we set N equals 64 for speed up.

\item{\textit{CNN Layer:}}
CNN layers is selected to get higher level representations. The convolutional layers target at selecting high level representations from the output of LSTM. The learned results will be feeded into the next dense layer.
We use two CNN blocks for feature extraction, each block contains filer and max pooling components., for the first filter component, we have 6 filters which size equals $ 5 \times 5 $ and stride equals 1 and for the first max pooling component which size equals $ 2 \times 2$ and stride equals 2. Hence after the first CNN block, we can get a $ 98 \times 30 \times 6$ vector. Then, we pass this vector to the second CNN block, in this block, we have 10 filters which size is $5 \times 3$ and stride is 3. Then we get a $ 32 \times 10 \times 10$ high dimensional features.

\item{\textit{Dense Layer:}}
After two CNN blocks, we can get a vector shaped as $ 32 \times 10 \times 10$, then we flat this vector into the shape of $ 3200\times 1$ and pass this vector to three fully-connected layers. Each layer contains 1000, 200, 5 neurons. After dense layer, we can get a $5 \times 1$ vector.

\item{\textit{Softmax Layer:}}
A softmax layer to output the predicted probabilities. The learned features by the previous CNN layers can be directly passed to a classifier like a softmax output layer to determine the likeliness. We utilize a 5-units softmax output layer to build a predictor on CSI amplitude and phase information.
\end{itemize}

\subsubsection{Online Testing}
We collect a large number of samples to find the relationship between number of people and CSI in the period of offline training and improve the performance of this end-to-end learning by our activity recognition model. At the online testing stage, we use this deep learning model to infer the number of people at present. If the predicted result is contradicted to the result obtained by activity recognition model. DeepCount will add this sample to retrain our deep learning model. We only retrain the parameters of the last dense layer in our deep learning model. The reason to update the parameters of last dense layer rather than whole layers' parameters is that the low-level features deep learning model extracted are similar. Hence, we can just retrain the last layer to get a better performance. At the same time, the time cost is negligible because we just retrain the single sample to the last layer. By automatically adjusting the model's parameters over a period of time, our recognition accuracy can up to \markedred{90\%}.

\begin{figure}
 \subfigure[Lab1]{
 
  \includegraphics[width=0.2\textwidth]{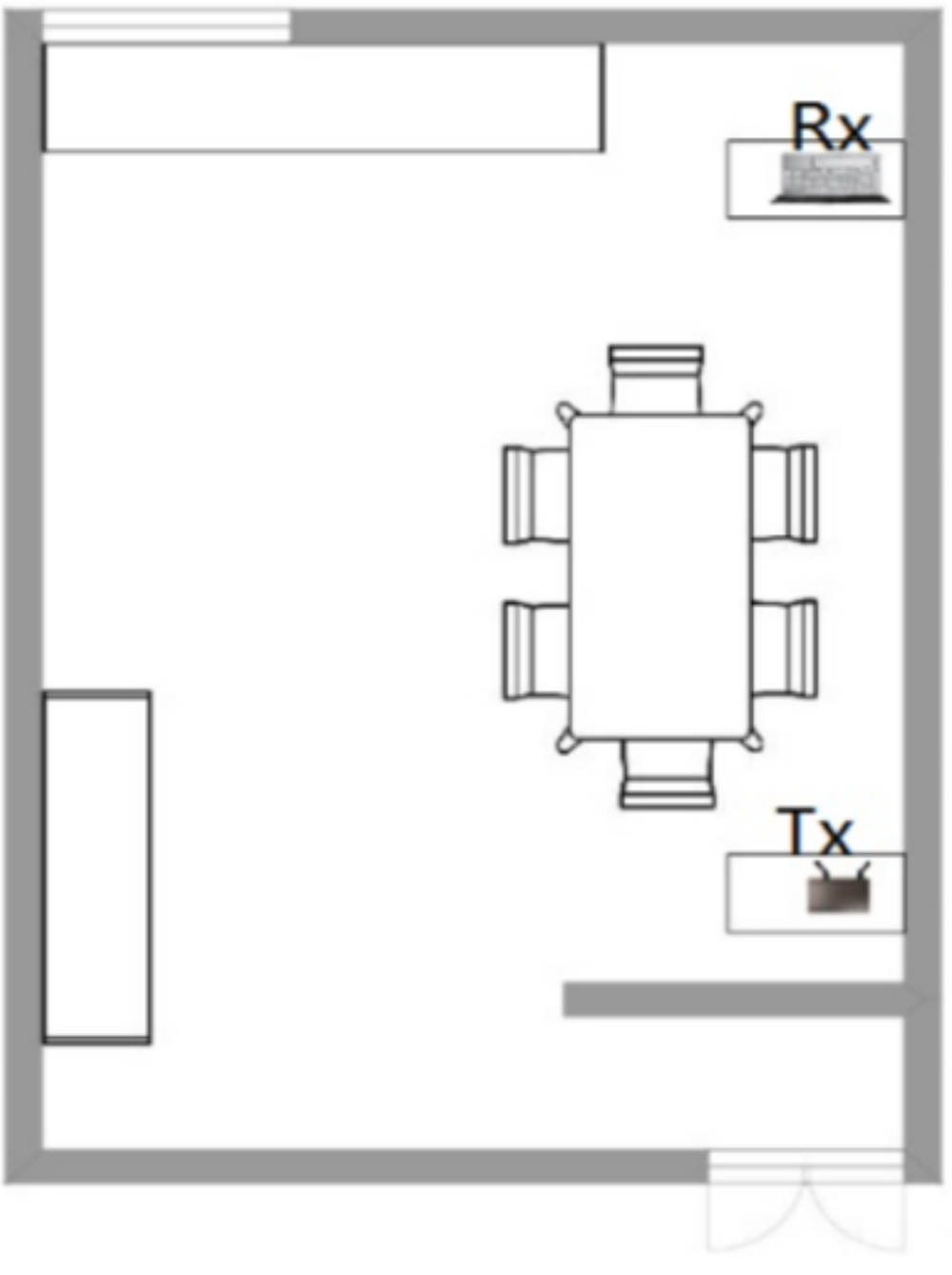}
  }
 \subfigure[Lab2]{
 
  \includegraphics[width=0.2\textwidth]{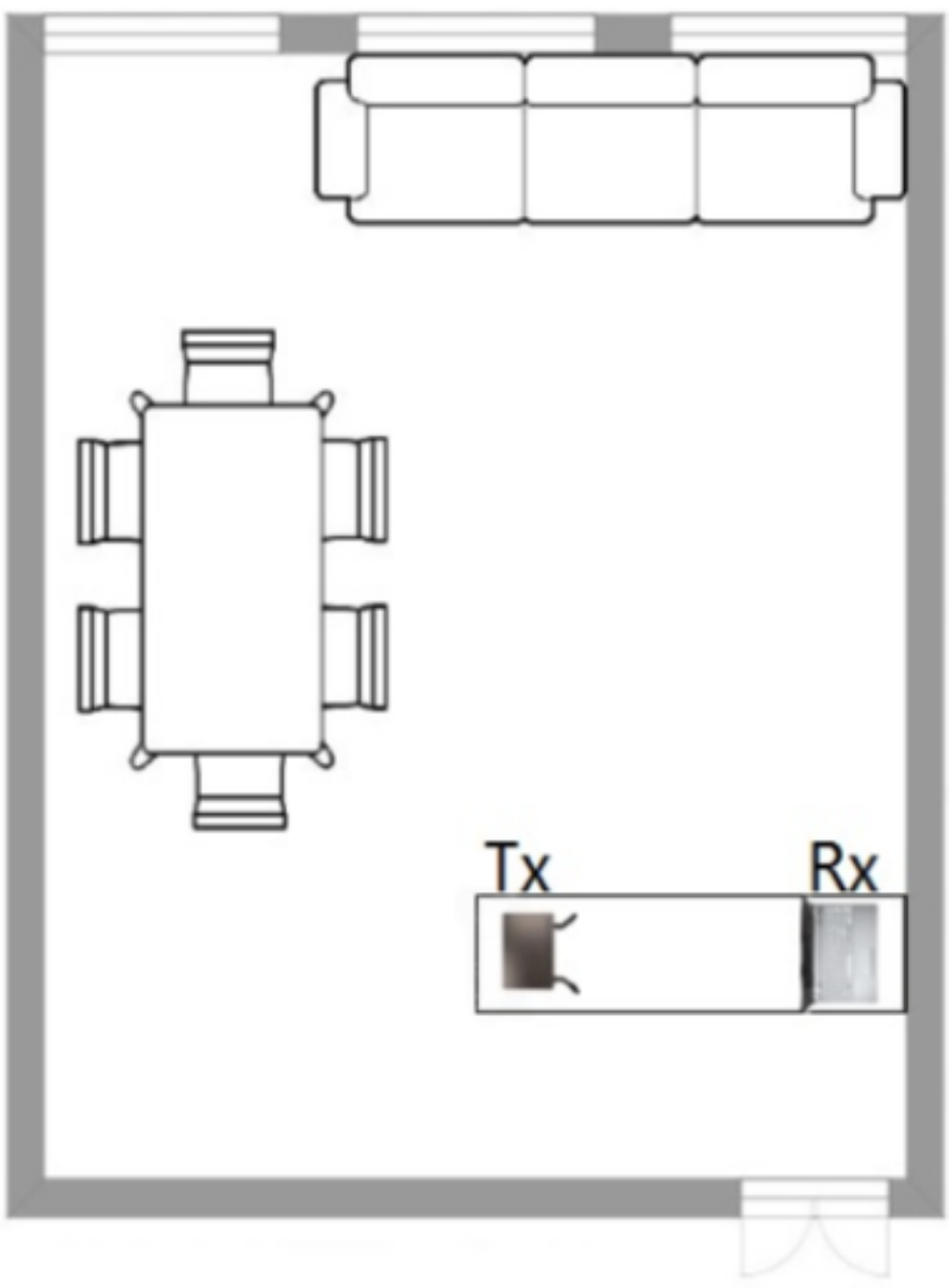}
  }
 \caption{Floor plans}
 \label{floor}
\end{figure}

\section{IMPLEMENTATION \& EVALUATION} \label{evaluation}
\subsection{Implementation}
In the experiment, the laptop used in our experiment is equipped with Ubuntu 12.04 operating system. In terms of hardware, our device is equipped with Intel 5300 NIC as receiver. We connected the laptop to a mini R1C wireless router with two antennas, using the router's cable as a transmitter.The receiver is equipped with three antennas and its firmware is modified to report CSI to the upper layer.
All experiments in this paper were carried out under the premise of a frequency band of 5 GHz and a channel bandwidth of 20 MHz. In order to make the wavelength short enough to ensure better resolution, we chose 5GHz instead of 2.4GHz, while 5GHz has more channels to reduce the possibility of inference. During the experiments, the transmitter  sends packets with a high rate of 1500 packets/second to the receiver continuously with Iperf tool. DeepCount acquires CSI measurements and processes it using Matlab, Python 3.6 and Tensorflow . We analysed the experimental data on the server with \markedred{two NVIDIA GeForce GTX 1080Ti GPU}. We use the cross entropy function as the loss function of the deep learning model to calculate the probability error in the classification while using the gradient descent optimizer (GDO / SGD) optimization algorithm as the optimizer to reduce the cost.
\subsection{Evaluation setup}
\begin{figure}
 
  \label{trainset}
  \includegraphics[width=0.45\textwidth]{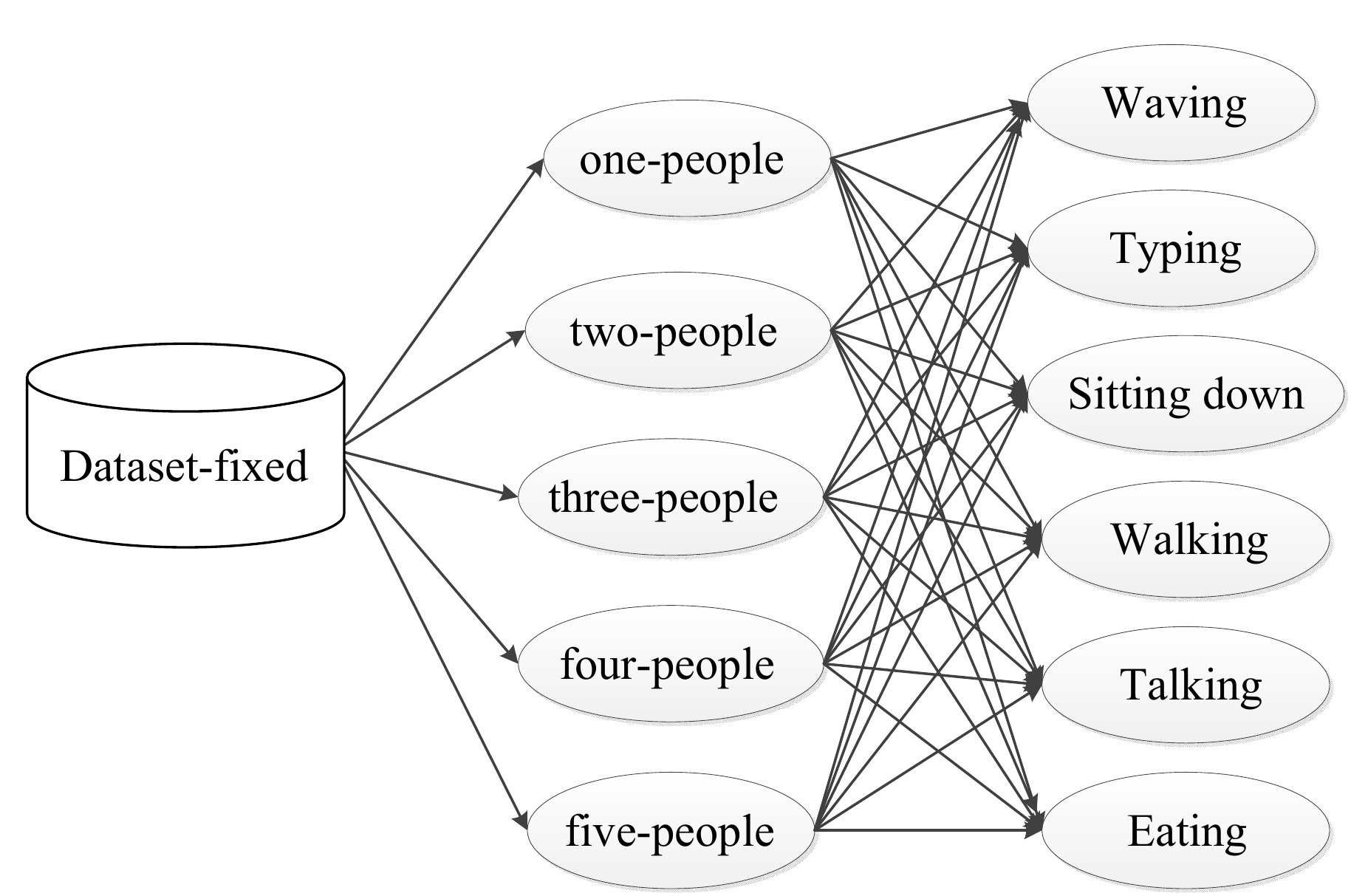}
  \caption{The structure of Dataset-fixed}
  \label{The structure of Dataset-fixed}
\end{figure}

Fig.\ref{floor} shows the training samples we collected  in lab1 and lab2. For our activity recognition model, We collected a total of 800 samples from 10 volunteers which include 7 male students and 3 female students for 8 different activities. Theses activities are listed in Table \ref{activitykind}, along with the number of samples for each activity. For our deep learning model, we divide our experimental process into three steps in order to prove the accuracy of our experiments: 
First, due to the uncertainty of human status and human location, we first let volunteers perform fixed activities at fixed locations.Volunteers do some fixed activities at designated locations or follow a designated route. Then we get the dataset-fixed which is shown in Fig.\ref{The structure of Dataset-fixed}. Second, we gradually relaxed the conditions. Volunteers are to choose actions which may be a combination of fixed actions above on a fixed position, the dataset we get is called dataset-semi. At last, we do not make any restrictions. Volunteers can engage in any activity anywhere in the room, the dataset is called dataset-open.For each sample, we collected 4 minutes of time series data, then a time window of size 200 splits the data and averages the amplitude and phase information for each time window to enlarge the training data set.  The splitted samples of different activities are shown in Table.\ref{Samples of dataset-fixed}.

\begin{table*}
\renewcommand\arraystretch{1.5}
	\centering
  \caption{The dataset of activity recognition model}
	\label{activitykind}
	\begin{tabular}{|c|c|c|c|c|c|c|c|c|c|}
		\hline
		 Activity: & Empty & Walking & Sitting down & Falling & Running & Entering into room & Leaving room & Waving \\ \hline 
		 Abbreviations: & E & W & S & F & R & O & L & A \\ \hline 
		 Samples: & 100 & 100 & 100 & 100 & 100 & 100 & 100 & 100 
		 \\ 		\hline
	\end{tabular}
\end{table*}

\begin{table}
	\centering
    \begin{tabular}	{|l|r|}
    \hline
        Activity & Samples \\ \hline
        Waving & 24741 \\ \hline
        Typing & 28565 \\ \hline
        Sitting down & 27108 \\ \hline
        Walking & 27537 \\ \hline
        Talking & 23580 \\ \hline
        Eating & 26802 \\ \hline

\end{tabular}
\caption{Samples of dataset-fixed}
\label{Samples of dataset-fixed}
\end{table}

 \subsection{\markedred{Baseline method} }
\subsubsection{Baseline method description }
\begin{figure}
\begin{minipage}[t]{0.5\textwidth}
\centering 
\includegraphics[width=0.9\textwidth]{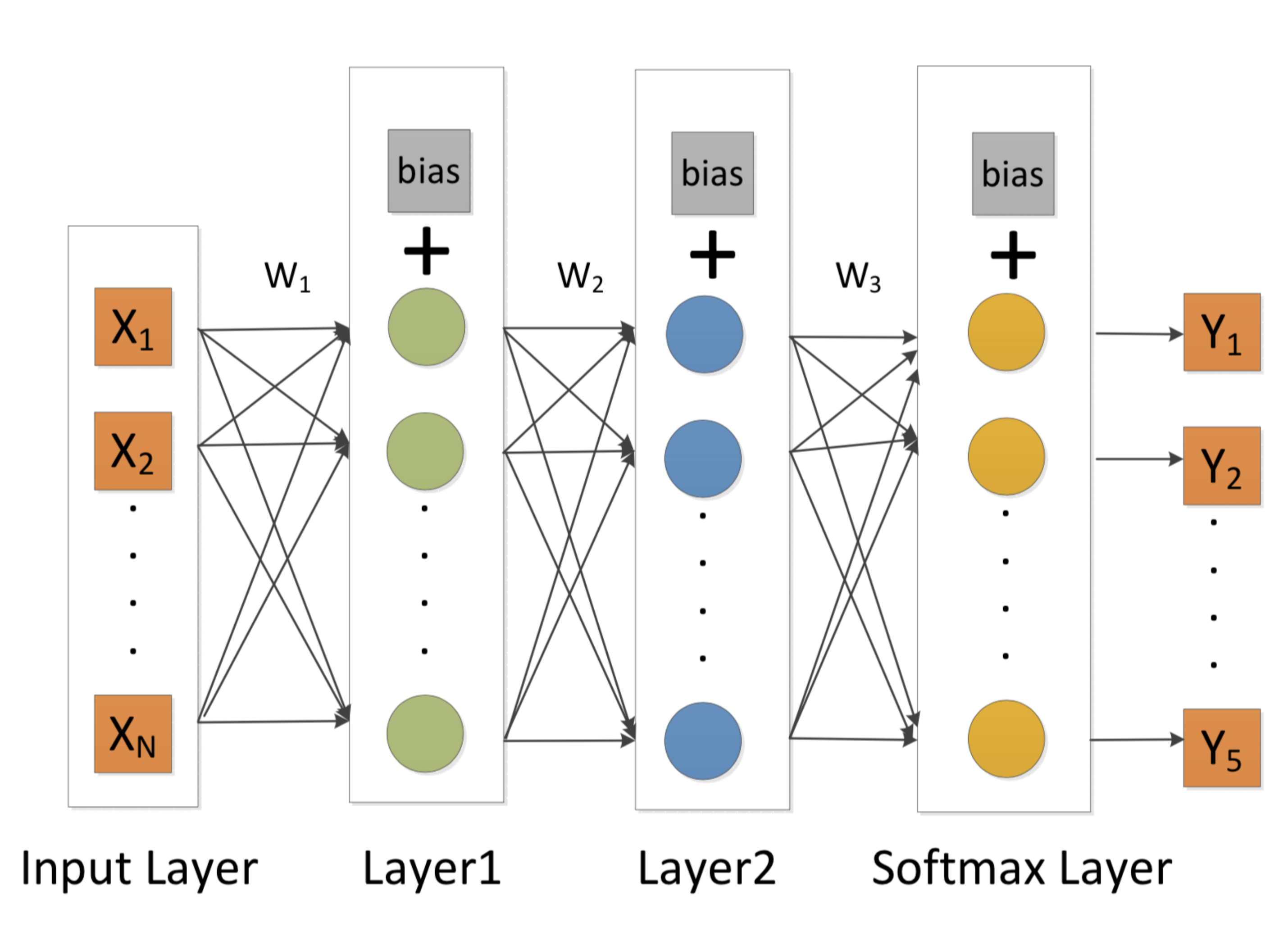} 
\caption{The architecture of baseline method network}
\label{priordeepnetwork}
\end{minipage} 
\end{figure}

 To examine the effectiveness of our method we add a baseline method into our experiment to form the control group.
 Fully Connected Back Propagation(FCBP) neural network which contains large parameters is suitable for crowd counting. Therefore ,we use FCBP neural network as the baseline method to compare our CNN-LSTM NetWork . The FCBP neural network we use has two hidden layers where each hidden layer consists of a different number of neurons. We use 300 neurons (denoted as Layer 1 nodes) on the first hidden layer and 100 neurons (denoted as Layer 2 nodes) on the second hidden layer.  The input layer is denoted as XN, where N equal 360 for the reason that CSI values both contain phase and amplitude information. The output has 5 classifications denoted as Y1 to Y5 to identify up to 5 people. Note that, this structure could easily extend to count more than 5 people by using more output nodes. Fig.\ref{priordeepnetwork} illustrates the structure of whole network.

 \subsubsection{\markedred{Experiment results of baseline method}}

\begin{figure*}
 \subfigure[The confusion matrix of baseline method for Dataset-fixed]{
  \label{fixed prior}
  \includegraphics[width=0.3\textwidth]{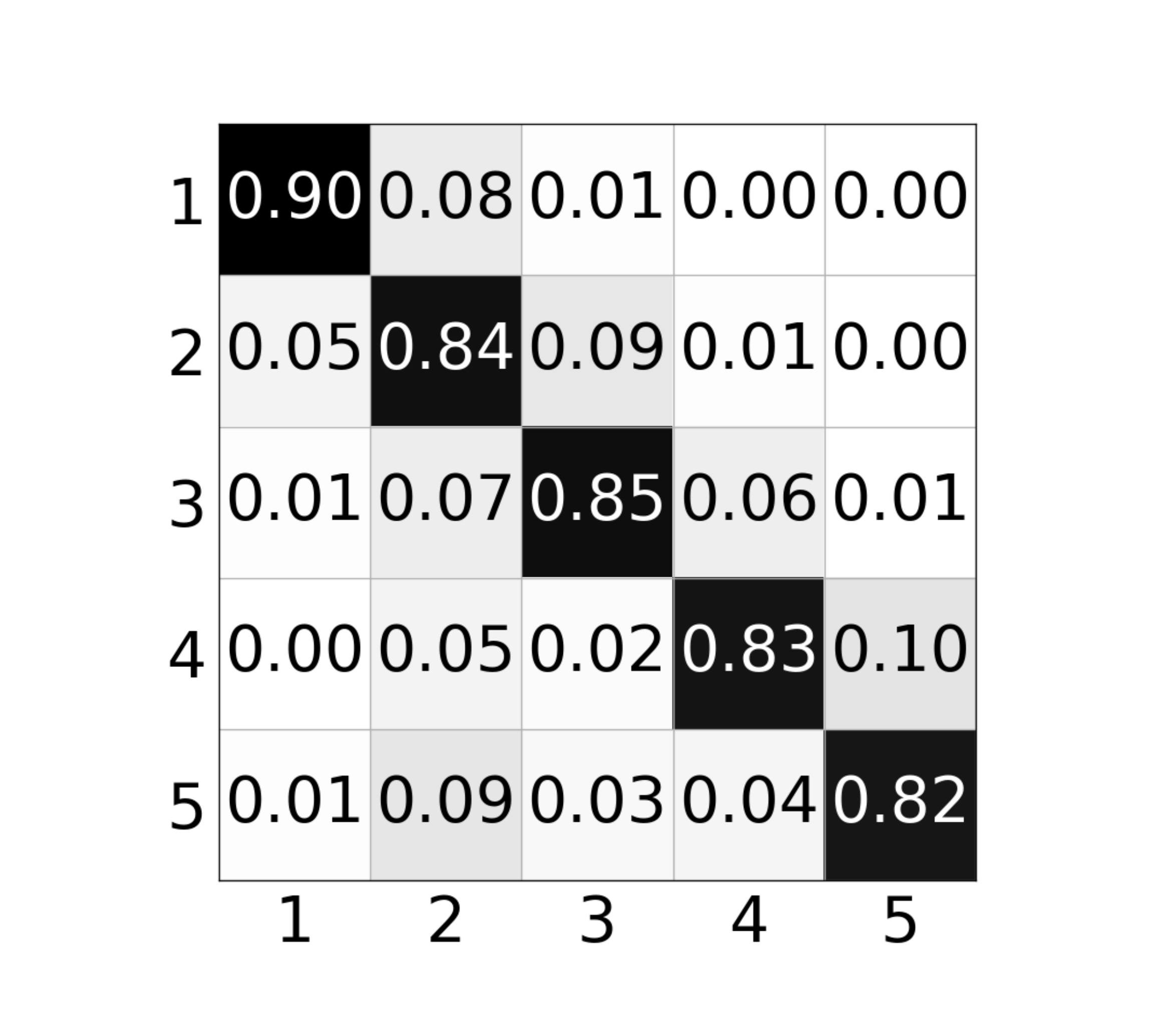}
  }
 \subfigure[The confusion matrix of baseline method for Dataset-semi]{
  \label{semi prior}
  \includegraphics[width=0.3\textwidth]{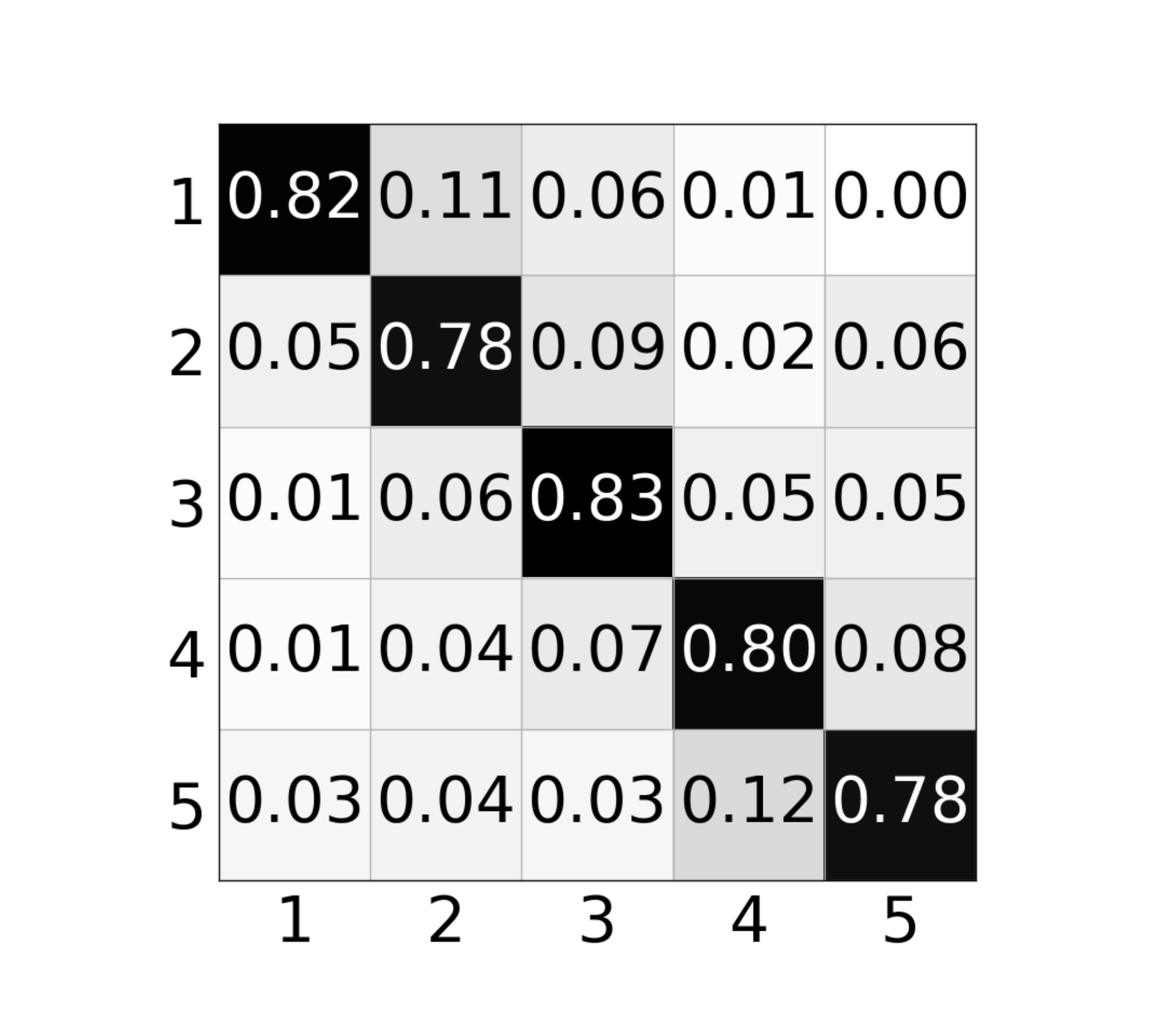}
  }
  \subfigure[The confusion matrix of baseline method for Dataset-open]{
  \label{open prior}
  \includegraphics[width=0.3\textwidth]{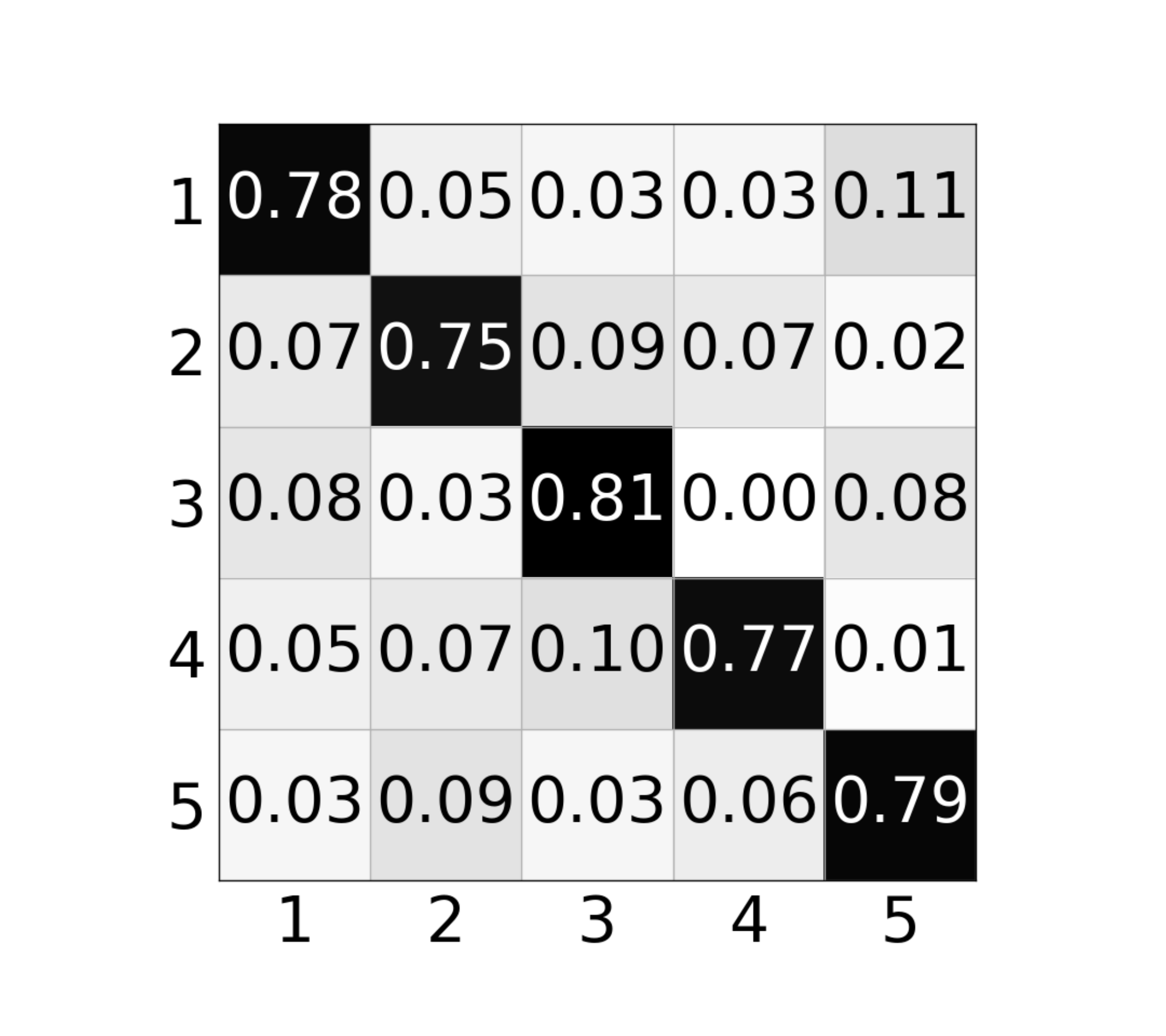}
  }
  \caption{\markedred{Confusion matrix of baseline method network}}
  \label{Confusion matrix of baseline method network}
\end{figure*}

\begin{figure}
\begin{minipage}[t]{0.5\textwidth}
\centering 
\includegraphics[width=0.9\textwidth]{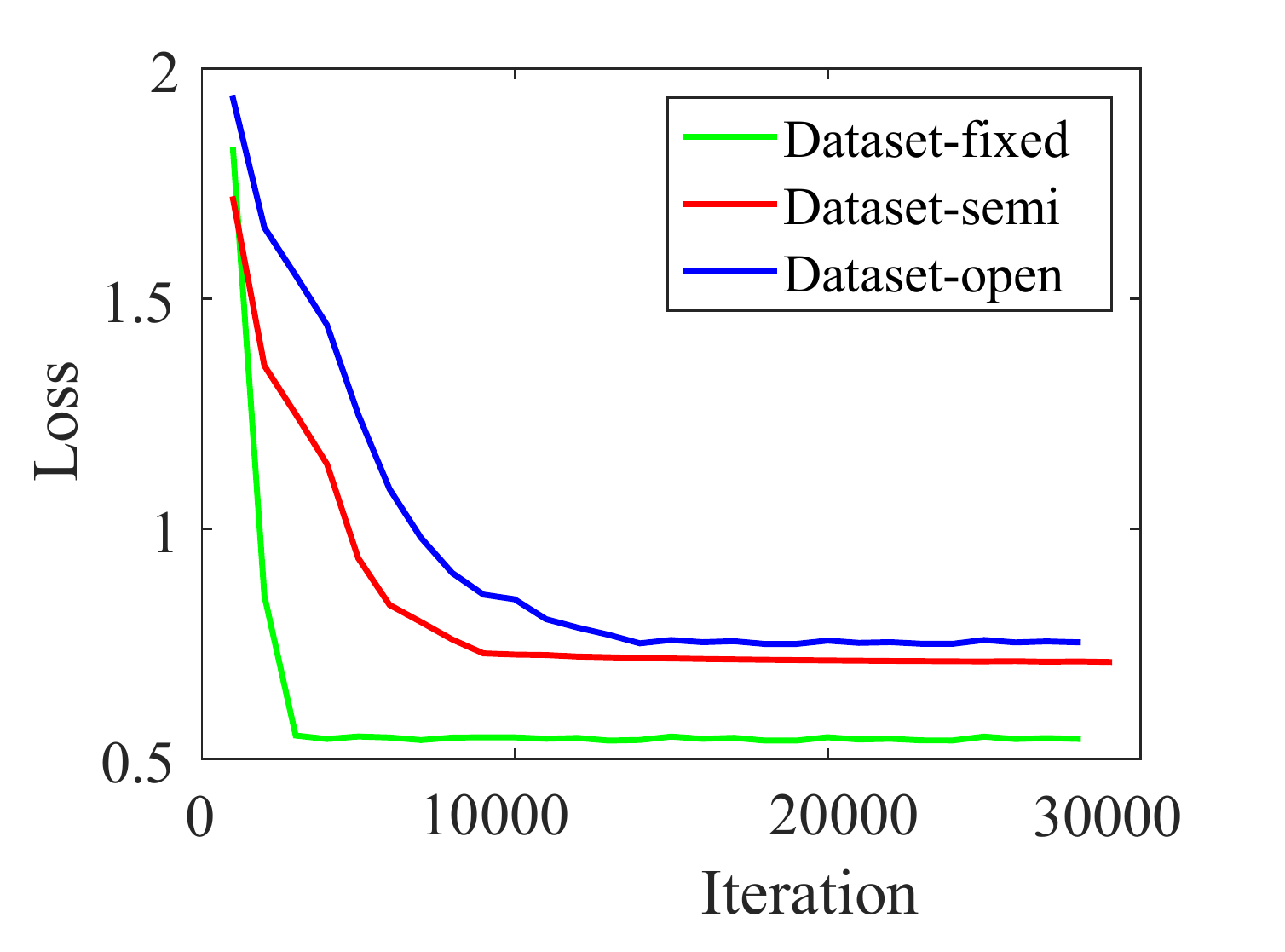} 
\caption{The loss curve of baseline method network}
\label{The loss curve of baseline method network}
\end{minipage} 
\end{figure}

The baseline method achieves an average accuracy of 88.8\%, 80.2\% and 78\% respectively across dataset-fixed, dataset-semi and dataset-open. Fig.\ref{Confusion matrix of baseline method network} plots the confusion matrix for different training dataset collected in our Lab1 and Lab2. Fig.\ref{The loss curve of baseline method network} shows the loss curve during the process of training. The loss function of dataset-fixed is converged after 3000 iterations while dataset-semi and dataset-open are converged even after 10000 iterations. This result implies that the data from dataset-semi and dataset-open are more diverse. Hence, the computation costs are relatively higher.

 \subsection{Experiment with CNN-LSTM}

\subsubsection{Parameters tuning}
As we all know, an effective network needs well-tuned parameters via extensive searching. The parameters include the number of layers, the number of units for each layer, the filter size and stride size for convolutional layer, and etc. The difficulty grows quickly with the complexity of a neural network. To balance the efficiency and performance of training process, we focus on tuning the major parameters that influence the performance most. We list down those parameters for each layer bellow.
\begin{itemize}
\item{\textit{LSTM Layer:}}
The performance of LSTM Layer is mostly sensitive to the maximum length of the input sequence, the number of LSTM cells, and dropout rate. LSTM with more cells means stronger information storage capacity. In our experiment, we use 1 layer of LSTM with 64 cells to remember long term dependencies. Also, We set the length of input sequences equals 200 for better performance and choose dropout rate equals 0.1.
\item{\textit{CNN Layer:}}
We use two CNN blocks to extract high level features. We tuned the filter and stride size for the convolutional layers, and the pool and stride size for the pooling layers. For the first block, we have the first CNN with $ 5 \times 5 $ filter, $1 \times $1 stride, followed by a max pooling with $2 \times $2 filter and $2 \times $2 stride, and the second block with $5 \times $3 filter, $ 3 \times $3 stride.
\item{\textit{Dense Layer:}}
Three fully-connected layers are followed by CNN blocks with 1000, 200, 5 neurons to reduce data dimension and fit the relationships between CSI and crowd counting.
\item{\textit{Other significant parameters:}}
In our experiment, we set the batch size equals 64 and learning rate equals \markedred{0.2,0.15,0,1 for the dataset-fixed,dataset-open and dataset-semi } to get the better performance.
\end{itemize}

\begin{figure}
 \subfigure[The accuracy of door switch among different activities]{
  \label{activity}
  \includegraphics[width=0.22\textwidth,height=0.12\textheight]{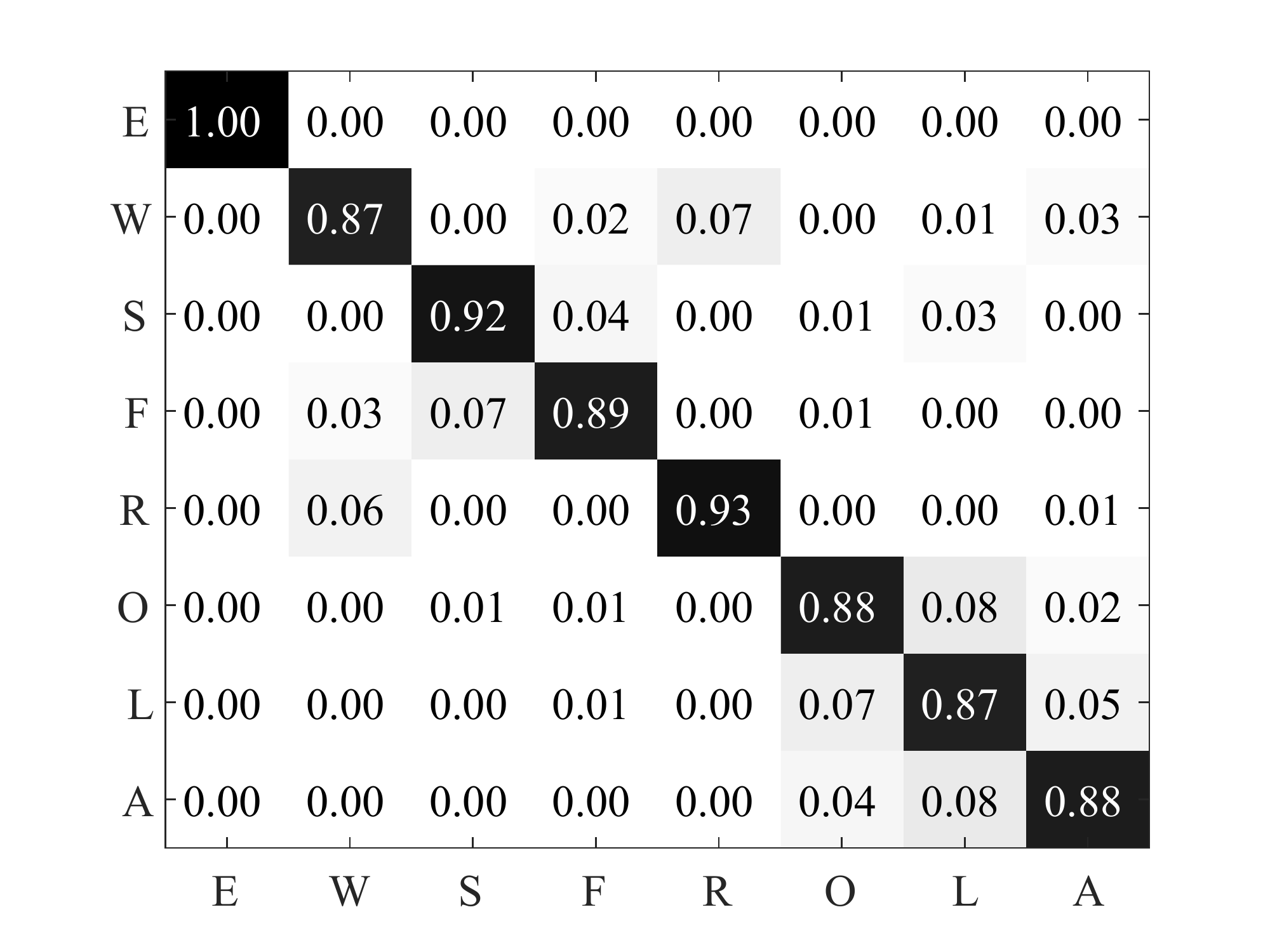}
  }
 \subfigure[The accuracy of door switch among different training set]{
  \label{train set size}
  \includegraphics[width=0.22\textwidth,height=0.12\textheight]{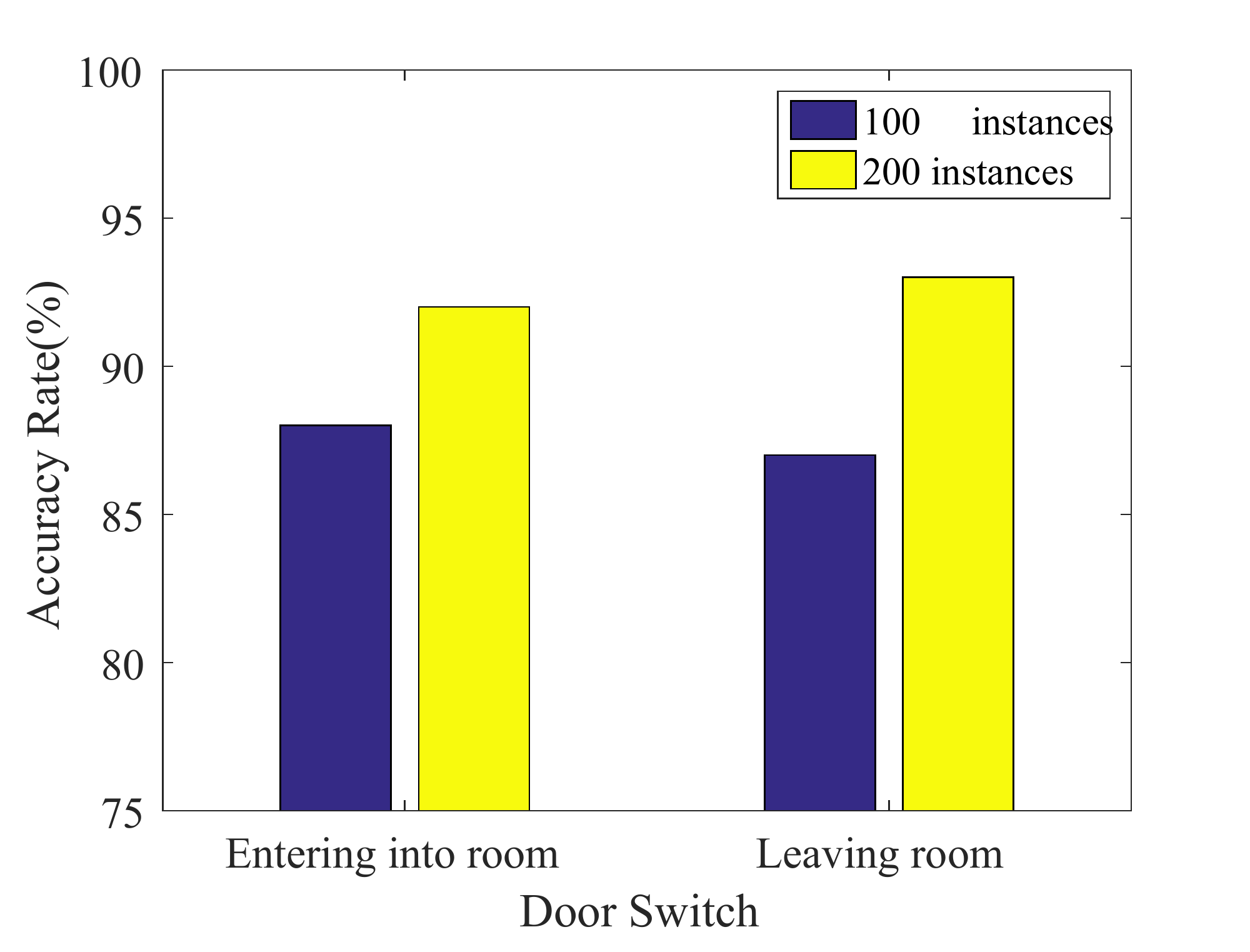}
  }
 \caption{Activity recognition model}
\end{figure}

\subsubsection{Overall performance}
\begin{figure*}
\subfigure[The confusion matrix of Dataset-fixed]{
  \label{fixed new}
  \includegraphics[width=0.3\textwidth]{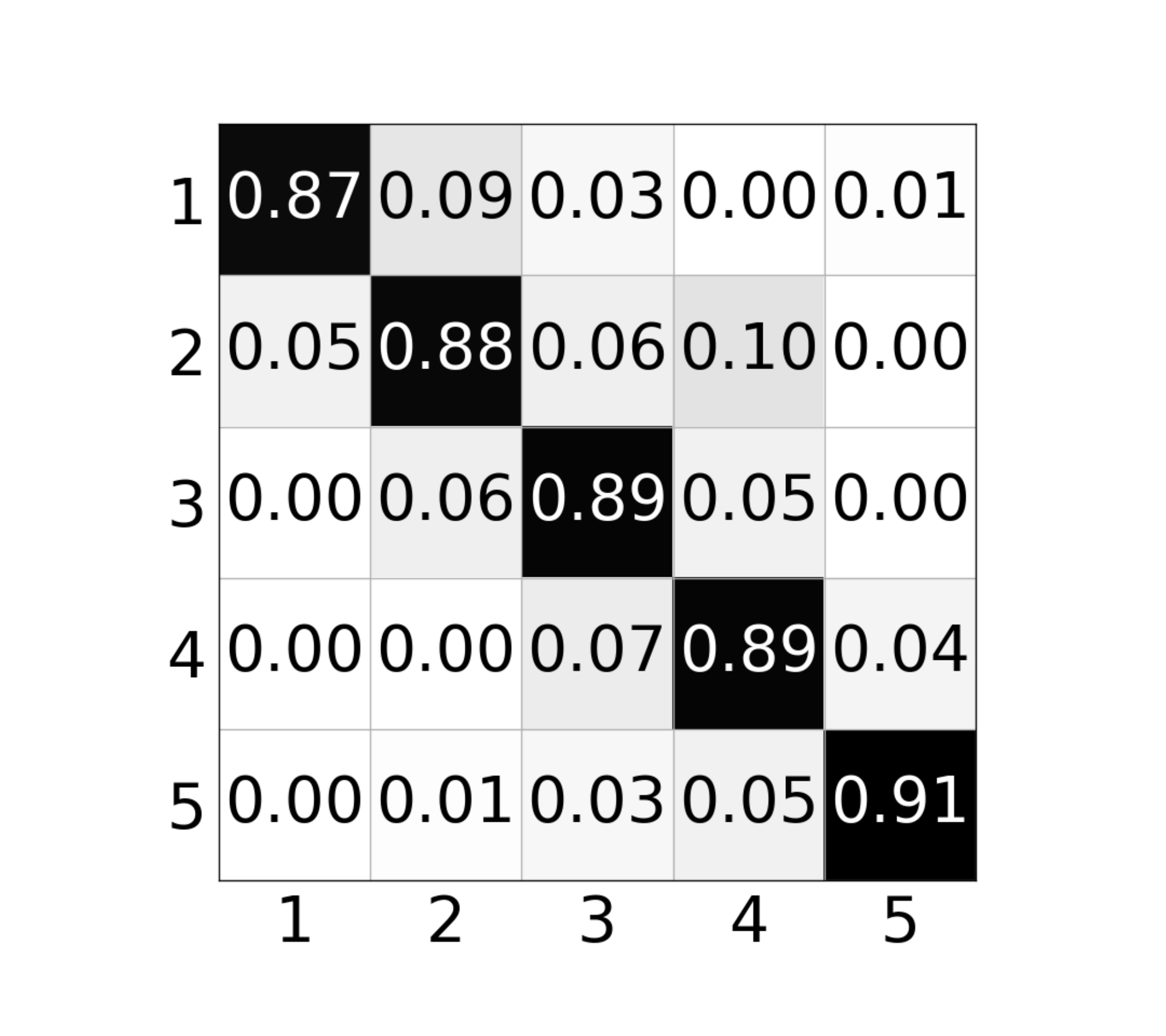}
  }
 \subfigure[The confusion matrix of Dataset-semi]{
  \label{semi new}
  \includegraphics[width=0.3\textwidth]{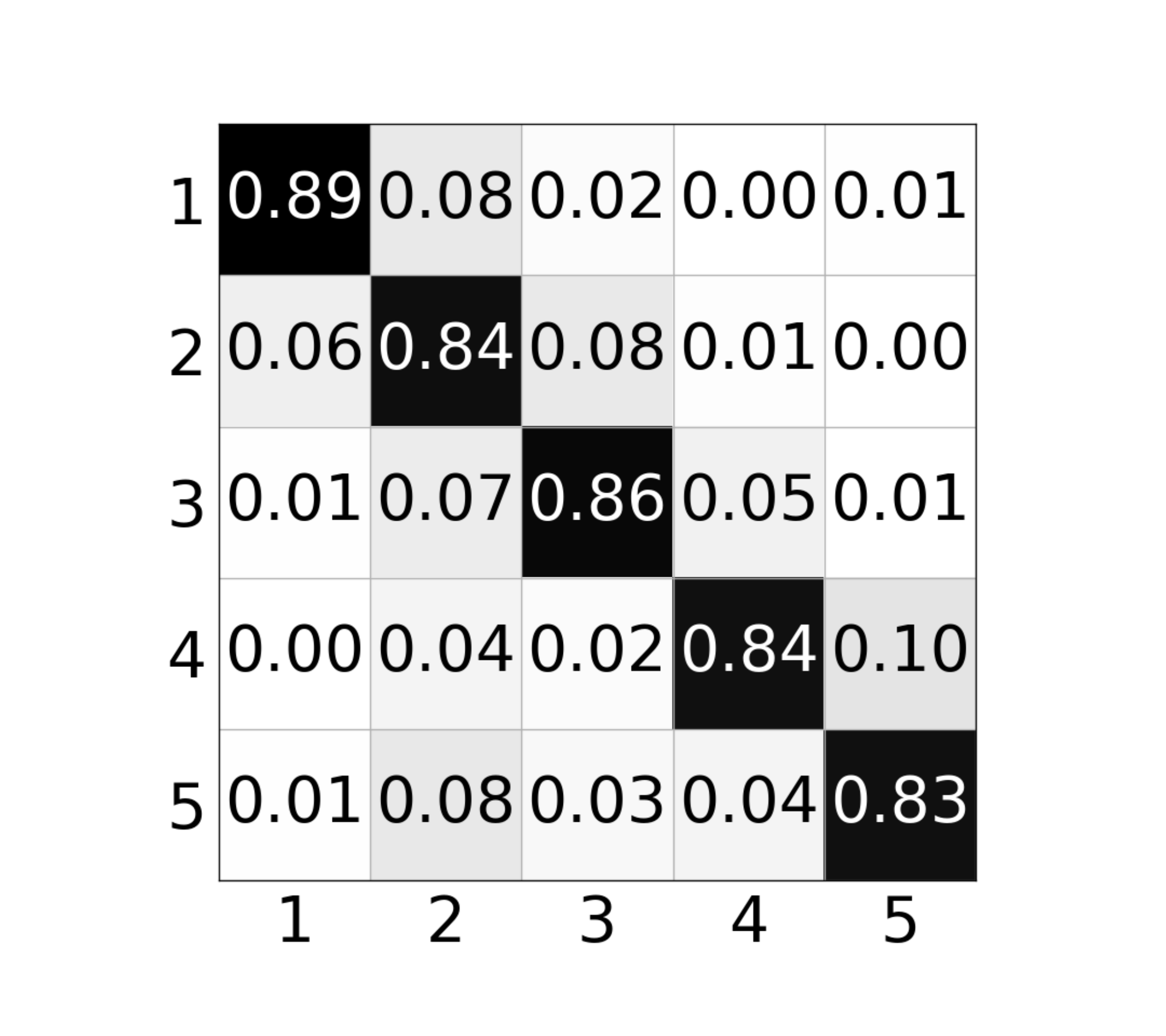}
  }
  \subfigure[The confusion matrix of Dataset-open]{
  \label{open new}
  \includegraphics[width=0.3\textwidth]{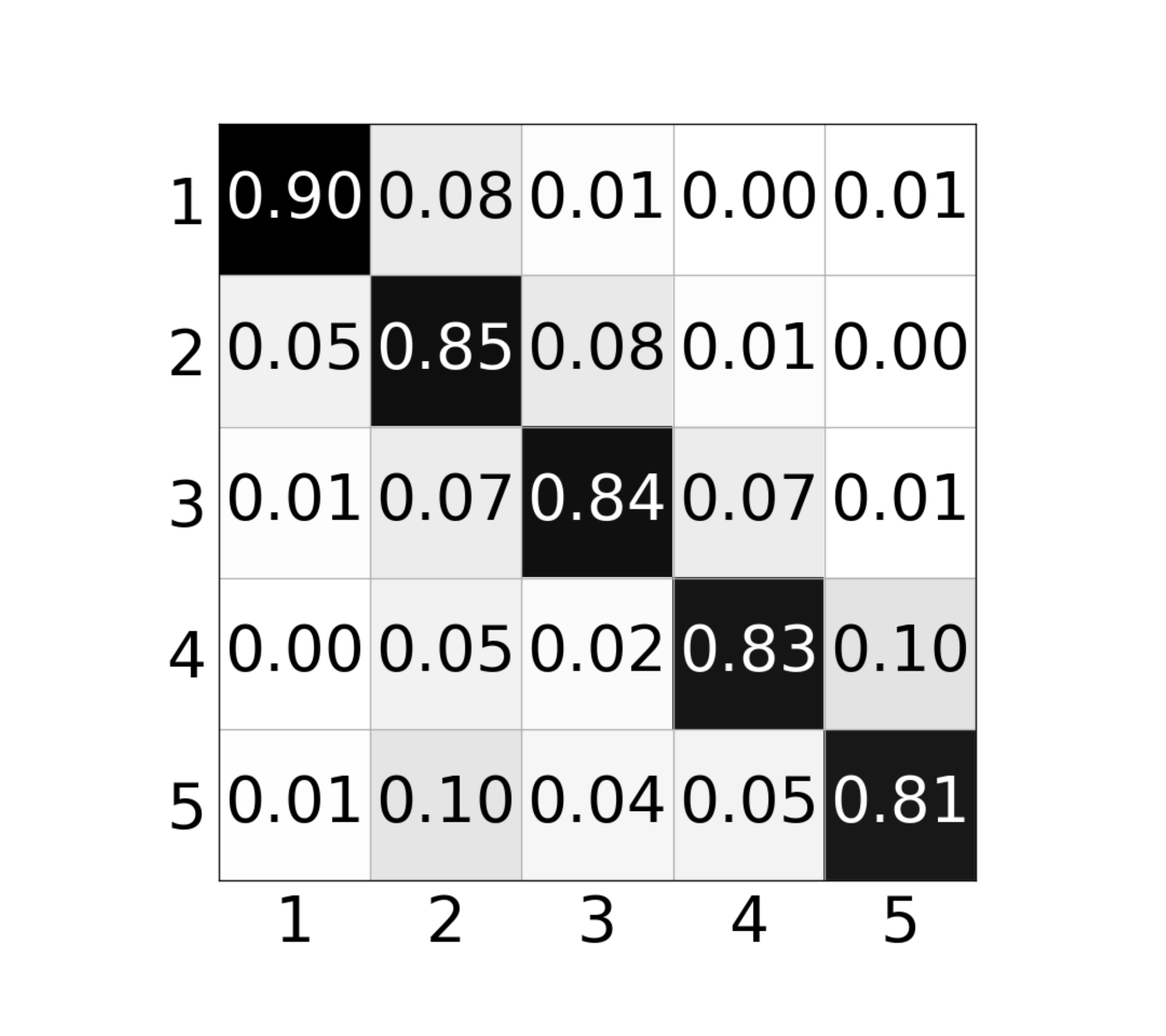}
  }
  \caption{\markedred{Confusion matrix of CNN-LSTM network}}
  \label{Confusion matrix of current network}
\end{figure*}

\begin{figure}
\begin{minipage}[t]{0.5\textwidth}
\centering 
\includegraphics[width=0.5\textwidth]{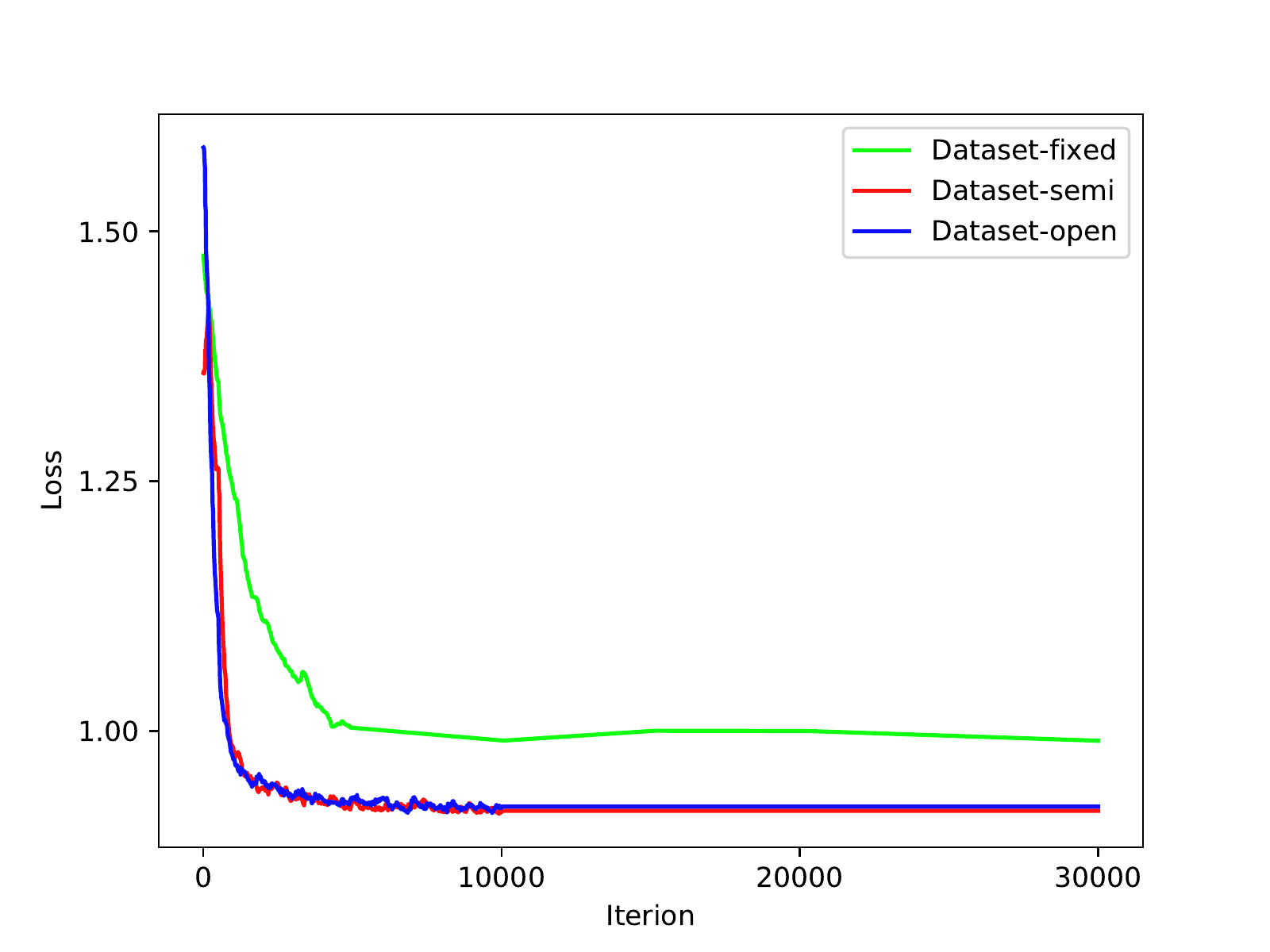} 
\caption{The loss curve of CNN-LSTM network}
\label{The loss curve of new methods network}
\end{minipage} 
\end{figure}

For activity recognition model, DeepCount takes the 80\% of samples in each class as the training set, the rest as the test set. For the training set, we use 10-fold cross validation to get optimal parameters for the activity model including the states in HMM. From Fig. \ref{activity}, the average accuracy of activity recognition is 89.14\% and there is a probability of 8\% to view the Entering into room as Leaving room and 7\% to view Leaving room as Entering into room. We observe that the accuracy of Entering into room and Leaving room is 88\% and 87\% among others activities. \markedred{For CNN-LSTM model, the DeepCount achieves an average accuracy of 88.8\%,85.2\%,and 85.2\% respectively across dataset-fixed, dataset-semi and dataset-open. Fig.\ref{Confusion matrix of current network} plots the confusion matrix for different training dataset in our Lab1 and Lab2. We can find that DeepCount's predicted results and real labels differ from no more than 2 people. Thus, we can see thatn our CNN-LSTM model fits the real environment well.} Due to the existence of multiple human states in the indoor environment, the data we collect cannot guarantee that all scenarios can be covered, which has a greater impact on our accuracy. On the other hand, 
despite such unfavorable factors, we can still achieve satisfactory results through deep learning methods. \markedred{Fig.\ref{The loss curve of new methods network} shows the loss curve during the process of training. We can find that the loss function of dataset-fixed and dataset-semi and dataset-open is converged after 3000,2500 and 2500 iterations.} We also use SVM as a control group and choose Gaussian kernel to train the data but the accuracy is less than 50\%. In order to determine whether neural network can distinguish more people, we also test scenario with 10 people in a completely open state. The accuracy achieves 75\%, which shows that neural network is powerful enough for crowd counting.

\subsection{\markedred{Comparison experiment result with baseline methods}}

\markedred{
For neural networks, the most direct benchmark of their performance is the accuracy and the convergence speed, so we made the following comparisons of the training results of baseline methods and CNN-LSTM.
}

\subsubsection{\markedred{Comparison of accuracy}}
\markedred{
From Fig.\ref{fixed prior} and Fig.\ref{fixed new} , we can find that when we fixed human activities and locations in the indoor environment, the accuracy of baseline method NetWork and CNN-LSTM is nearly equal. But Fig.\ref{open prior} and Fig.\ref{open new} shows that for the open-dataset, the 85.2\% accuracy of CNN-LSTM is better than the 78\% accuracy of baseline method NetWork. Meanwhile, From Fig.\ref{semi prior} and Fig.\ref{semi new} , we can find that the 80.2\% accuracy of CNN-LSTM is better than the 85.2\% accuracy of baseline method NetWork for the dataset of semi. Thus, from the Fig.\ref{Confusion matrix of baseline method network} and the Fig.\ref{Confusion matrix of current network}, we can get the conclusion that the experiment accuracy of CNN-LSTM is higher and more stable than baseline method.
}

\subsubsection{\markedred{Comparison of converged iterations}}
\markedred{
The Fig.\ref{The loss curve of baseline method network}
and Fig.\ref{The loss curve of new methods network} shows that the different converged iterations of baseline method and CNN-LSTM . We can find for the dataset-fixed, the converged iterations of baseline method and CNN-LSTM is 3000, so the baseline method and CNN-LSTM has the same efficiency about converged iterations for dataset-fixed. But for the dataset-open and dataset-semi, the baseline method converged after 10000 iterations but the CNN-LSTM method only need 2500 iterations. With above performance results, we can come to the conclusion that the CNN-LSTM Network is converged faster than baseline method.
}

\subsection{\markedred{Experiment of Different Components}}
\subsubsection{Impacts of Activity Training Set Size}
To examine the effect of training set for different activity samples on the accuracy of DeepCount, we increase the number of samples from 100 to 200 per activity. Fig. \ref{train set size} plots the accuracy with different training set size. We observe that the average accuracy of door switch increased from 87.5\% to 92.5\%. When we increase the number of samples, we also improve the computational cost of DeepCount.

\subsubsection{Impact of time window}
\begin{figure}[htb]
\centering 
\begin{minipage}[t]{0.26\textwidth}
\includegraphics[width=0.9\textwidth]{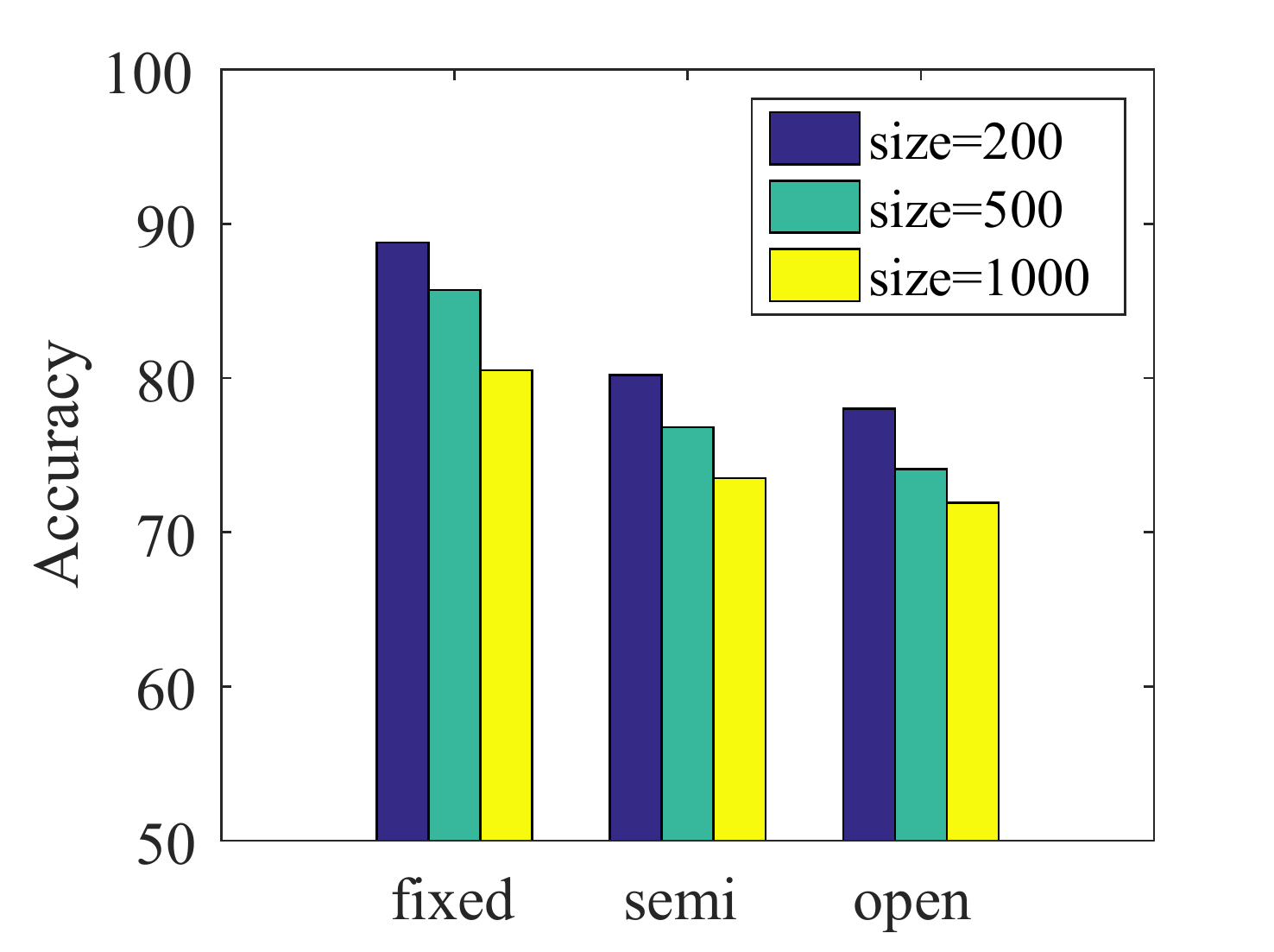} 
\caption{The impact of the length of time window}
\label{differentsamples}
\end{minipage} 
\end{figure}

To enlarge the dataset ,we use the length of 200 as the time window size to split the train data . In order to compare the effect of different time window sizes on recognition accuracy, our time window size was set to 500 and 1000 as comparative experiments.The accuracy is shown in the Fig.\ref{differentsamples}. We found that when the time window size is equal to 200, the recognition accuracy reaches the highest point. Although the time window becomes larger, the number of samples that we need to train can be reduced. However, under the premise of a sampling rate of 1500 packets/second, it is difficult to ensure that the CSI value remains relatively stable for the duration of the time window. Therefore, we cannot increase the time window size in order to reduce the sample of training.

\subsubsection{Impact of preprocessing}
\begin{figure}
 \subfigure[The accuracy of data with or without preprocessing ]{
  \label{preprocess}
  \includegraphics[width=0.22\textwidth,height=0.12\textheight]{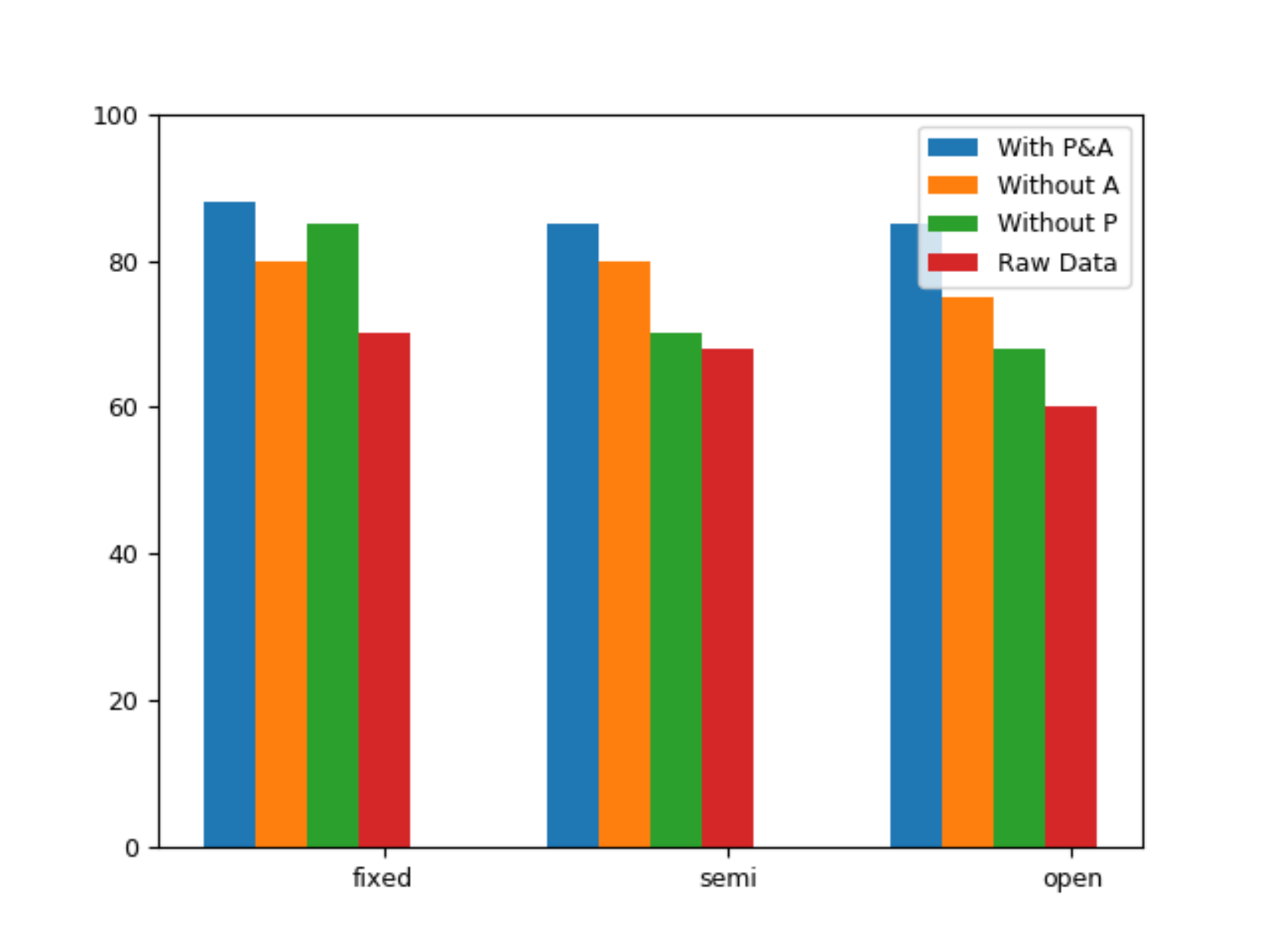}
  }
 \subfigure[The accuracy of different features]{
  \label{differentfeatures}
  \includegraphics[width=0.22\textwidth,height=0.12\textheight]{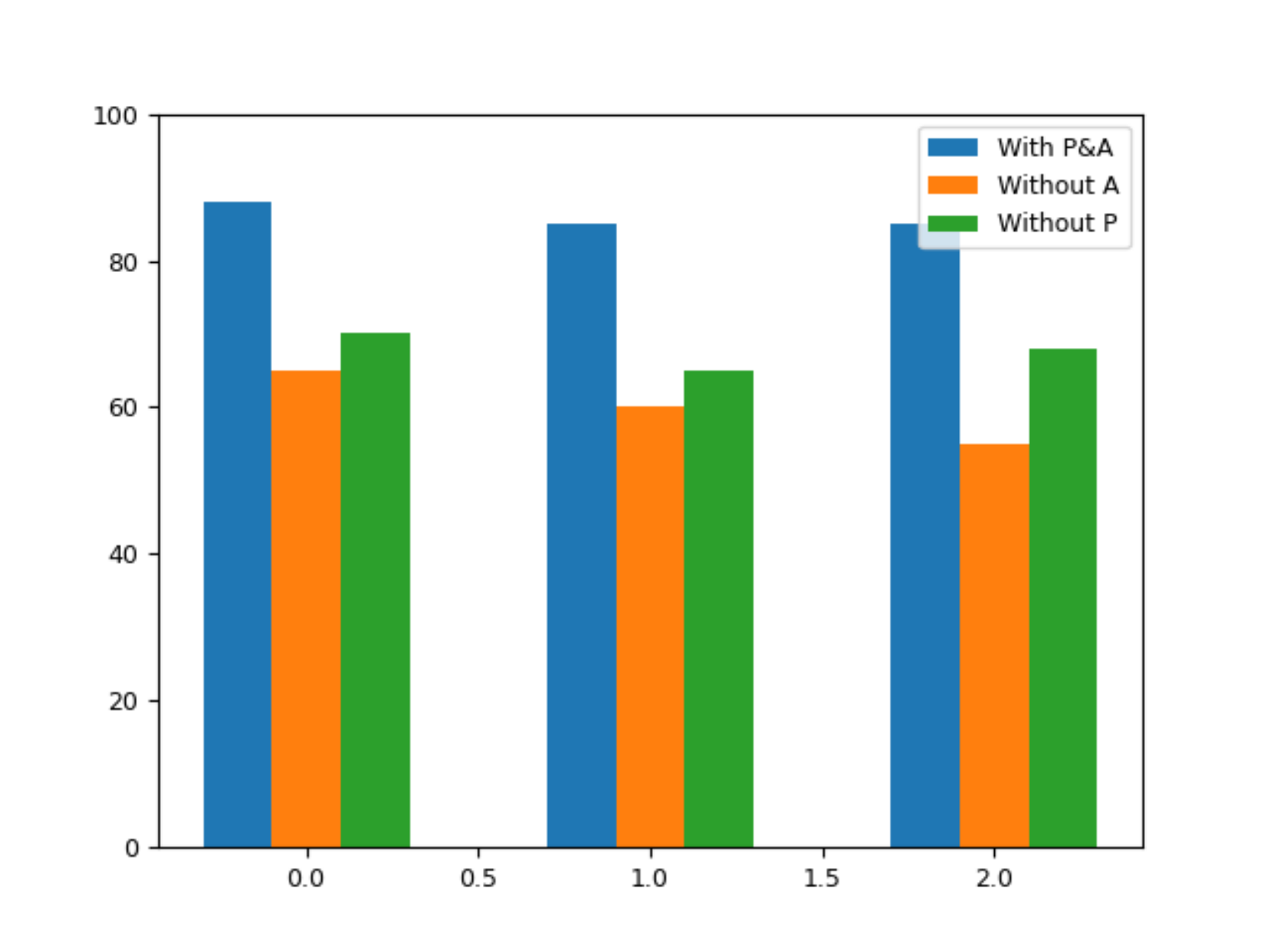}
  }
 \caption{Impact of preprocessing}
\end{figure}

The main influencing factors of preprocessing are  phase sanitization and amplitude noise removal. Therefore, it is necessary to verify whether these processes have a large impact on the accuracy of DeepCount. We also want to know what effect we will have on the accuracy of DeepCount if we only train amplitude information or only train phase information. From Fig.\ref{preprocess} where legend With P\&A denotes the data with phase sanitization and amplitude noise removal, legend Without A denotes the data without amplitude noise removal, legend Without P denotes the data without phase sanitization and legend Raw Data denotes the data without preprocess, we find that if without phase sanitization and amplitude noise removal,the accuracy of DeepCount will drops .This shows that eliminating some significant data noise will significantly improve the performance of the neural network.
This shows that eliminating significant noise in the data will significantly improve the performance of the neural network. As is shown in Fig.\ref{differentfeatures}, where legend With P\&A denotes both phase and amplitude information as features, legend Without P denotes we only use amplitude as features and legend Without A denotes phase information as features. We found that the recognition rate using both amplitude and phase information is higher than using only one of the information. It shows that increasing the number of features within a certain range can improve the performance of DeepCount, and that the amplitude and phase information are directly related to the number of people.

\subsubsection{Impact of amendment mechanism}
By analyzing the results of the activity recognition model, we can find that HMM can effectively distinguish door switch from other activities. Hence, we can use this model to judge the accuracy of deep learning model. If the predicted result is inconsistent with the activity recognition model, we add this sample to retrain the last layer parameters. Owing to retrain the last layer of single sample, the time cost is acceptable. \markedred{By this mechanism, we can eventually improve the recognition accuracy from 82.3\% to 87\% on the baseline method . Further, the recognition accuracy of the CNN-LSTM network can be improved from 86.4\% to 90\%.}

\subsubsection{Compared with Electronic Frog Eye}
DeepCount is much different from the Electronic Frog Eye \cite{ref17}. First, FCC temped to find a monotonic relationship between CSI variations and the number of people. Based on these, they proposed an algorithm called Dilatation-based crowd profiling to character the variations. However, we find that the relationships are far more complicated owing to the uncertainty states in the indoor environment and we can not only use the amplitude variations to determine the number of people in our experiment. Second, the Grey theory used to predict the number of people is not reliable. As we all know, Grey theory is to establish prediction model to make a vague description of the development of things by a small amount of incomplete information. However, crowd counting is a relatively precise task and in many cases, we should exactly know the number of people. At last, FCC estimated the number of people with multiple devices, however, they do not solve the problem well about signal interference and synchronization problems between these devices. In our experiment, we utilize the powerful learning ability of neural networks to solve this problem, and at the same time, we add activity recognition model to amend the wrong predicted results. The experimental results show that we can get the accuracy up to \markedred{90\%} and our results are more reliable.

\section{CONCLUSIONS}\label{conclusion}
In this paper, we present DeepCount, a novel system which solve multi-human environmental sensing problems use the deep learning approach with WiFi signals. To the best of our knowledge, it is the first solution to use neural networks as a crowd counting solution. To further improve the performance of DeepCount, we add an online learning mechanism to get better results. DeepCount also explores why deep learning can solve this complex problem. Preliminary results show that compared to the traditional classification algorithm such as SVM, DeepCount can achieve an average recognition accuracy of \markedred{86.4\%} for up to 5 people, showing a higher recognition rate. With the help of activity recognition model, we can get the accuracy up to \markedred{90\%}. Our approach can show acceptable accuracy in the context of complex changes in the indoor environment, which means our approach works fairly robust.
Although the deep learning approaches require huge amount of samples to fit this complex function and in our experiment we collect massive samples to characterize correlations between CSI and crowd counting. In theory, if we can take into account enough circumstances in the indoor environment and take these as samples to build a robust model, we can reuse the model for the same environment. We plan to explore how to improve the maximum distinguished number of people as our future work.

\section*{Acknowledgment}
This work is supported in part by the National Key R\&D Program of China under Grant 2017YFB0802300, the National Science Foundation of China under Grant (No.61602238, 61672283), the key project of Jiangsu Research Program Grant( BK20160805), the China Postdoctoral Science
Foundation (No. 2016M590451).

\bibliographystyle{unsrt}

\end{document}